\documentclass[lettersize,journal]{IEEEtran}
\usepackage{amsmath,amsfonts}
\usepackage{algorithmic}
\usepackage{array}
\usepackage[caption=false,font=normalsize,labelfont=sf,textfont=sf]{subfig}
\usepackage{textcomp}
\usepackage{stfloats}
\usepackage{pifont}
\usepackage{booktabs}
\usepackage{xcolor}
\definecolor{my_orange}{RGB}{255,192,0}
\definecolor{my_blue}{RGB}{0,176,240}
\definecolor{my_purple}{RGB}{112,48,160}
\definecolor{my_green}{RGB}{112,173,71}

\usepackage{url}
\usepackage{verbatim}
\usepackage{multirow}
\usepackage{float}
\usepackage{graphicx}
\def\BibTeX{{\rm B\kern-.05em{\sc i\kern-.025em b}\kern-.08em
    T\kern-.1667em\lower.7ex\hbox{E}\kern-.125emX}}
\usepackage{balance}
\begin{document}
\title{Towards Student Actions in Classroom Scenes: New Dataset and Baseline}
\author{Zhuolin Tan, Chenqiang Gao, Anyong Qin, Ruixin Chen, Tiecheng Song, Feng Yang, and Deyu Meng
\thanks{Manuscript received Month, Year; revised Month, Year. This work is supported in part by the National Key R\&D Program of China  (Grant No.2022YFA1004100), and in part by the National Natural Science Foundation of China (Grant No. 62176035, 62201111), the Shenzhen Fundamental Research Program (Grant No. JCYJ20240813151216022). \textit{(Corresponding author: Chenqiang Gao, Anyong Qin.)}}
\IEEEcompsocitemizethanks{
\IEEEcompsocthanksitem Zhuolin Tan, Anyong Qin, Ruixin Chen, Tiecheng Song, and Feng Yang are with the School of Communications and Information Engineering, Chongqing University of Posts and Telecommunications, Chongqing 400065, China (e-mail: tanzhuolin98@gmail.com; qinay@cqupt.edu.cn).
\IEEEcompsocthanksitem Chenqiang Gao is with the School of Intelligent Systems Engineering, Shenzhen Campus of Sun Yat-sen University, Shenzhen, Guangdong 518107, P.R. China, and also with the School of Communications and Information Engineering, Chongqing University of Posts and Telecommunications, Chongqing 400065, China (e-mail: gaochq6@mail.sysu.edu.cn).
\IEEEcompsocthanksitem Deyu Meng is with the School of Mathematics and Statistics and the Ministry of Education Key Laboratory of Intelligent Networks and Network Security, Xi’an Jiaotong University, Xi’an, Shaanxi 710049, China (e-mail: dymeng@mail.xjtu.edu.cn).
}}

\maketitle

\begin{abstract}
Analyzing student actions is an important and challenging task in educational research. Existing efforts have been hampered by the lack of accessible datasets to capture the nuanced action dynamics in classrooms. In this paper, we present a new multi-label \textit{Student Action Video} (SAV) dataset, specifically designed for action detection in classroom settings. The SAV dataset consists of 4,324 carefully trimmed video clips from 758 different classrooms, annotated with 15 distinct student actions. Compared to existing action detection datasets, the SAV dataset stands out by providing a wide range of real classroom scenarios, high-quality video data, and unique challenges, including subtle movement differences, dense object engagement, significant scale differences, varied shooting angles, and visual occlusion. These complexities introduce new opportunities and challenges to advance action detection methods. 
To benchmark this, we propose a novel baseline method based on a visual transformer, designed to enhance attention to key local details within small and dense object regions. Our method demonstrates excellent performance with a mean Average Precision (mAP) of 67.9\% and 27.4\% on the SAV and AVA datasets, respectively. 
This paper not only provides the dataset but also calls for further research into AI-driven educational tools that may transform teaching methodologies and learning outcomes. The code and dataset are released at https://github.com/Ritatanz/SAV.
\end{abstract}


\section{Introduction}
\IEEEPARstart{I}{n} the realm of education, understanding student actions in the classroom is important for assessing the effectiveness of teaching methods \cite{li2023dynamic,dillon2019effects,duong2019brief,royer2019systematic}. 
Traditional observational techniques, while insightful, are often limited by subjective interpretations and fail to provide a comprehensive view of student engagement and learning actions. The integration of computer vision technology offers a more objective and detailed perspective by capturing and analyzing student body language, behavioral changes, and interactions. This technological approach enables educators to accurately assess student focus, participation levels, and interest in the content, thereby facilitating customized adjustments to teaching strategies to meet diverse learning needs better. Researchers have demonstrated that the presence of computer technology in classrooms is essential \cite{dimitriadou2023critical,heflin2017impact,mcknight2016teaching}. 
It not only avoids interfering with student actions, but also actively increases participation and enhances the educational experience. This underscores the importance of integrating and sustaining technology in educational environments to promote a more interactive and effective learning atmosphere.

Action recognition \cite{wang2021tdn,yang2020temporal,song2023joints,bertasius2021space,arnab2021vivit} and detection \cite{zhao2017temporal,li2020spatiotemproal,shi2023tridet,liu2022end,vahdani2022deep} are critical tasks in the fields of computer vision and artificial intelligence, which can be applied in various domains such as surveillance and sports analysis. Widely acknowledged action recognition datasets such as HMDB \cite{kuehne2011hmdb}, UCF101 \cite{soomro2012ucf101}, Kinetics \cite{carreira2017quo}, and action detection dataset AVA \cite{gu2018ava} have significantly advanced these fields. 
However, the development of behavior analysis techniques in educational settings is hampered by the lack of large-scale, publicly available video datasets of classroom behaviors.
Most existing benchmarks focus on daily and sports scenarios, leaving a gap in data availability for real educational environments. This gap severely impedes the exploration of algorithm performance and model robustness in educational contexts, posing a key challenge for researchers in the field.

\begin{figure}[!t]
\centering
\centerline{\includegraphics[width=8.5cm]{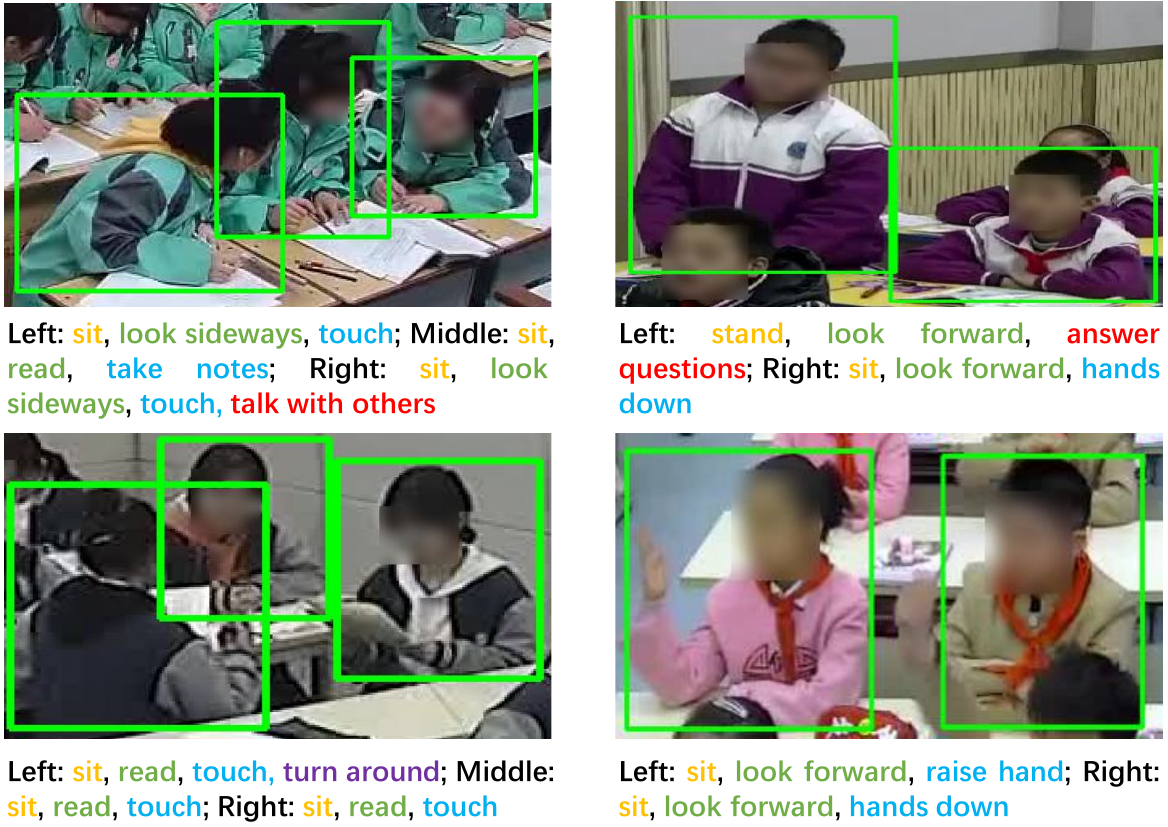}}
\caption{\color{black}The bounding boxes and action annotations in sample frames of our dataset. Each frame is cropped for zooming in to show the annotations better. Each person has a postural action 
(in \color{orange} orange\color{black}), 
a sight action (in \color{my_green} green\color{black}), 
person-object interactions (in \color{my_blue} blue\color{black}), 
body-motion actions (in \color{my_purple} purple\color{black}), 
and person-person interactions (in \color{red} red\color{black}) annotated when they occur. Note that only keyframes are shown here, and accurate annotation of actions requires temporal context.}
\label{introduction}
\end{figure}

In this paper, we present a multi-label \textit{Student Action Video} (SAV) dataset in complex classroom scenes. The dataset contains 4,324 video clips with 15 actions annotated, including a variety of classroom scenarios. Through these videos, we can provide a comprehensive perspective to analyze the diverse actions that students spontaneously exhibit in various classroom environments, thereby truly reflecting their learning and interaction patterns.
These actions are categorized into five typical types (see Fig. \ref{introduction}): \color{orange} postural action \color{black} (orange text), 
\color{my_green} sight action \color{black} (green text), 
\color{my_blue} person-object interaction \color{black} (blue text), 
\color{my_purple} body-motion action \color{black} (purple text), 
and \color{red} person-person interaction \color{black} (red text), including 15 specific actions such as sitting, standing, looking forward, reading, raising hands, taking notes, etc. In our analysis of the captured raw video, we noticed that a large portion of the video is mainly the explanation on the blackboard without any human activity. Meanwhile, we found after extensive observation that a video length of three seconds is ideal for capturing key information about both long-duration actions (e.g., postural and sight actions) and brief interactions (e.g., person-object and person-person interactions). This duration captures brief actions in their entirety and reduces the information redundancy of long-duration actions that are static or have small deformations. Therefore, to focus on student actions, we segment the original video into three-second clips, selectively retaining the clips containing student activity.

The SAV dataset has the following three principal characteristics. Firstly, it contains a wide range of realistic classroom scenarios, collected from surveillance videos across 758 classrooms, covering a diverse range of curricula and educational stages. This variety ensures an authentic representation of student actions, making the dataset highly valuable for both academic research and practical applications. Secondly, the dataset has high-quality video resolutions, primarily 720P and 1080P, which greatly exceeds the 320×400 resolution of datasets such as AVA  \cite{gu2018ava}. This high resolution allows for the preservation of detailed visual information, enhancing the utility of the dataset. 
Lastly, the inherent complexity of real classroom scenarios presents numerous challenges for existing algorithms, such as subtle action variations, dense objects, significant scale differences, varied filming angles, and visual occlusions. Addressing these challenges is essential for developing robust models that can perform accurately in educational settings.



Existing Visual Transformers (ViTs) \cite{dosovitskiy2020image} perform well in capturing global information in visual data. However, they often struggle to effectively capture critical local details when confronted with small and dense objects in SAV. 
To overcome this limitation, we propose a new baseline method based on an improved ViT architecture. The method combines a Local Relation Aggregator (LRA) that focuses on small-scale local interactions. This ensures that local dynamics within small object regions are fully captured. In addition, we propose a Window Enhanced Attention (WEA) module. It identifies local window regions with the strongest feature responses in each token and performs spatio-temporal attention on these regions to focus on discriminative tiny objects. By leveraging the ability of ViT to capture global relations and the attention of LRA and WEA to local details, our method aims to enhance the understanding and analysis of complex classroom scenes.
Our baseline method achieves a mean Average Precision (mAP) of 67.90\% on the SAV dataset and 27.4\% on the AVA dataset, demonstrating competitive performance against current state-of-the-art methods. Furthermore, our study explores the critical role of temporal information in the SAV dataset, the difficulty of identifying different actions, and the impact of LRA and WEA on performance.


The main contributions of this work can be summarized as follows:
\begin{enumerate}
    \item We develop a Student Action Video (SAV) dataset in classroom scenes, which comprises 4,324 video clips and 15 categories. The dataset has diverse and real classroom scenes, high-quality video data, and numerous challenges. 
    \item We propose a novel baseline method specifically for dealing with small and dense objects in SAV. The method utilizes a ViT architecture enhanced by a Local Relation Aggregator module and a Window Enhanced Attention module, which improves the performance of action detection for student actions by focusing on spatio-temporal features both globally and locally.
    \item We conduct extensive experiments to evaluate the performance of the newly developed baseline method against existing state-of-the-art action detection methods. The results show that the proposed method obtains competitive results on the SAV and AVA datasets.
\end{enumerate}

\section{Related work}
\subsection{Action recognition and detection datasets}

Early action recognition datasets mainly focus on action classification, such as the KTH \cite{schuldt2004recognizing}, UCF-101 \cite{soomro2012ucf101} and HMDB \cite{kuehne2011hmdb}. These datasets contain short manually trimmed clips to capture individual daily actions and are suitable for introductory studies in action recognition.
Large-scale datasets such as Sports-1M \cite{karpathy2014large} and YouTube-8M \cite{abu2016youtube} cover an extensive array of sports and daily activities, facilitating the training of complex models to accommodate real-world variability. Kinetics \cite{carreira2017quo} expanded action diversity and is often used as pre-training for downstream tasks. Something-Something \cite{goyal2017something} focused on gestures and object interactions, offering valuable insights into object-related actions.
For action detection datasets, UCF Sports \cite{rodriguez2008action}, UCF101-24 \cite{soomro2012ucf101}, and J-HMDB \cite{jhuang2013towards} focused on evaluating single and coarse-grained action categories in short videos. AVA \cite{gu2018ava} and AVA-Kinetics \cite{li2020ava} included more sparse annotations of daily life actions, to reduce manual labor for annotation. MultiSports \cite{li2021multisports} is a recent work focusing on action detection in multi-person sports scenarios, providing additional challenges such as multi-person interactions and complex backgrounds in sports.

The era of big data has notably increased the accessibility of classroom video data, promoting in-depth analysis of classroom action. Numerous researchers have developed specialized datasets for classroom scenes. For example, recent works \cite{zheng2020intelligent,zhou2022classroom,zhao2023bitnet}
captured different numbers of images from local schools or publicly accessible educational videos. They focused on some coarse-grained actions for recognition, e.g., raising hands, standing, and sleeping. Zheng et al. \cite{zhou2022classroom} focused on human skeletons to recognize specific student actions. Sun et al. \cite{sun2021student} compiled classroom videos of different subjects from 11 classrooms. This dataset is used not only for action recognition, but also for temporal detection and automatic behavioral descriptions in video captions. These image- or video-based classroom datasets provide the basis for the practical application of complex action analysis in educational settings.

Although researchers have conducted the aforementioned studies on classroom student actions, these datasets have certain limitations.
First, these datasets are not comprehensive in their focus on student actions. Studies \cite{zheng2020intelligent} and \cite{zhou2022classroom} focused only on the simple and typical categories of student actions. Meanwhile, the coarse action annotations in \cite{zhao2023bitnet} and \cite{sun2021student} failed to capture the fine-grained distinctions necessary for accurate behavioral analysis in real-world educational settings.
Second, these datasets typically annotated only one action per individual at any given time, ignoring the possibility of concurrent actions. For example, a student may be annotated as ``looking around" without being aware of accompanying actions such as ``discussing". This leads to an incomplete description of student actions.
Finally, these datasets are typically limited to a single type of instructional setting and do not include a variety of educational stages, curricula, and population densities. This limitation hampers the generalization of models trained on these data.
Therefore, there is an urgent need to construct a complete, fine-grained, multi-scenario video dataset of student actions in the classroom, to quickly adapt to the complex open teaching environments of various types of schools in practical applications.


\subsection{Video action detection}

Video action detection aims to localize and recognize action instances within video frames. It plays a crucial role in applications such as surveillance, sports analytics, and video summary generation, where understanding dynamic activities in videos is essential.
The existing deep learning-based video action detection methods can be mainly divided into CNN-based \cite{liu2019multi,chen2020afnet,gan2022temporal} and transformer-based methods \cite{qing2023mar,fan2021multiscale,li2022mvitv2}.

In the CNN-based methods, YOWO \cite{kopuklu2019you} processed the spatial and temporal information of the video separately through spatial and temporal streams, which were finally merged at the classification layer to make a joint decision.
SlowFast \cite{feichtenhofer2019slowfast} used a slow path at a low frame rate and a fast path at a high frame rate to capture spatial semantic information and subtle temporal changes, respectively. It exchanged information through a specific fusion strategy.
In recent years, with the development of deep learning techniques, Vision Transformer (ViT) \cite{dosovitskiy2020image} has begun to show its powerful performance in the field of image processing. 
ViT split the image into small patches and treated these patches as sequences, capturing the global dependencies of the image, which draws on the Transformer architecture of Natural Language Processing \cite{vaswani2017attention}.
Subsequently, in order to better adapt to the characteristics of video data, Fan et al. \cite{fan2021multiscale} proposed the Multiscale Vision transformer (MViT), which extends ViTs by introducing a multiscale processing mechanism that enables the model to capture spatio-temporal features at different scales in different network depths. This is particularly important for understanding dynamic video content.
MViTv2 \cite{li2022mvitv2} performed shift-invariant positional embeddings using decomposed location distances to inject position information in the Transformer block, and compensated for the effect of pooling strides in attention computation through a residual pooling connection. The method further optimizes the architecture based on MViT to improve performance and computational efficiency.

Self-supervised learning (SSL) \cite{hendrycks2019using} is a machine learning framework designed to mine representational properties as supervisory information from unlabeled data, without the need for external supervisory signals. SSL is typically performed in a two-stage framework consisting of pre-training on unlabeled datasets, and fine-tuning for downstream tasks. Among them, MAE (Masked Autoencoder) \cite{he2022masked} is a typical example of applying self-supervised learning. MAE masked a portion of the input image, leaving only some of the pixels visible, which is then fed to a decoder transformer to reconstruct the masked portion of the original image.
The high masking rate of MAE improves the efficiency of the model, and enables the model to learn deep visual features. Similarly, VideoMAE extended this concept to video data \cite{tong2022videomae}. It reconstructs spatio-temporal information through random tube masking, encouraging the extraction of more efficient video representations during pre-training. VideoMAE is a data-efficient learner for self-supervised video pre-training. It outperforms existing fully-supervised methods such as SlowFast and MViT. 
However, VideoMAE utilized a vanilla ViT to capture global dependencies, which may not be sufficient in complex scenes with small and dense objects. Local detail information is crucial for accurately recognizing and understanding tiny objects in these complex scenes.
In contrast, our improved ViT incorporates the Local Relation Aggregator (LRA) and Window Enhanced Attention (WEA) modules, designed to effectively capture local token relationships in small and dense object regions. These enhancements significantly improve the ability of the model to learn discriminative local details, leading to higher accuracy in handling fine-grained tasks.


\section{The Student Action Video Dataset}
Our SAV dataset aims to introduce a new challenging benchmark to the field of action detection. The benchmark is specialized for classroom scenarios and defines fine-grained action categories. Section A describes our annotation process. The statistics and characteristics of SAV are detailed in Sections B and C.
\subsection{Data collection}

\textbf{Action Vocabulary Generation.} 
Based on the work \cite{mccoy2020gen} on student actions in the classroom, this study considers the fine-grained movements of the heads, eyes, mouths, upper bodies, and hands of students. We categorize these actions into five distinct types: pose, sight, person-object, body-motion, and person-person interactions. Postural actions include standing and sitting. Sight actions involve looking forward, looking around, and reading. Person-object interactions include flipping a book, raising a hand, taking notes, clapping, hands down, and touching. Body motion comprises bending and turning around. Person-person interactions include talking with others and answering questions. There is less variation between actions that belong to the same type. For instance, holding a pen without moving is categorized as touching, whereas using a pen to write on paper is classified as taking notes. These categories provide a comprehensive overview of the current learning state of students, thus allowing educators to conduct more precise analyses of student actions.


\textbf{Data preparation.} The data for this study comes from videos publicly posted on online educational platforms, which are freely accessible to anyone. In this way, we collect 758 videos from different schools in China and mainly cover kindergarten, primary, and middle school, which are the key stages of education.
The number of participants in each classroom varies from 4 to 68. After the initial data collection, we perform data cleaning, keeping only high-resolution videos in 720P and 1080P formats. Considering that some videos lack visible human subjects, we segment the videos into three-second clips. We then manually filter these clips to retain those containing visible student activities. These selection processes are critical for enhancing the suitability of the data for behavioral analysis and ensuring its overall quality.

\textbf{Person bounding box annotation.} 
Bounding boxes are used to locate action instances. Given the high object density in classroom videos and the minimal movement of human subjects, we have optimized the annotation process for efficiency. Following AVA \cite{gu2018ava}, we select the middle frame of each three-second video clip as the key frame for annotation. 
To ensure the quality of the annotations, we implement a hybrid strategy. First, we fine-tune a pre-trained Faster R-CNN person detector \cite{ren2016faster} using a subset of manually labeled classroom data. The model then performs initial bounding box detection on the videos. Subsequently, the annotator manually checks and corrects missed or inaccurately bounding boxes, thereby confirming the accuracy of the annotation.

\textbf{Action annotation.} All actions performed by each participant are labeled by a team of crowd-sourced annotators, resulting in a multi-label dataset. Each participant is assigned between one and five possible labels. To enhance the consistency and accuracy of the annotations, we develop a comprehensive labeling manual and conduct multiple rounds of cross-validation. These rigorous processes ensure the quality of the dataset and provide a solid foundation for subsequent analysis of student actions.
Regarding privacy protection, all annotators are required to sign a strict non-disclosure agreement before participating in the annotation process. This agreement explicitly prohibits them from distributing, sharing, or modifying any of the original video content. In addition, strict privacy measures are implemented throughout the database construction process. Personal information, such as names, addresses, and other sensitive details, is either removed or anonymized to safeguard the identity and security of the individuals in the dataset.

\subsection{Data statistics}



\begin{table*}[t]
\centering
\caption{Comparison of statistics between existing action detection datasets and our SAV. $Tube$ denotes that annotation has spatial localization and temporal boundary; $Frame$ indicates that annotation has spatial localization. $Multi-Person$ indicates that more than one person is annotated in the video; $Multi-label$ indicates that one person has multiple labels.}
\begin{tabular}{ccccccccccc}
\toprule
\toprule
Datasets       & J-HMDB \cite{jhuang2013towards}     & UCF101-24 \cite{soomro2012ucf101}          & AVA \cite{gu2018ava}   & Sun et al. \cite{sun2021student} & MultiSports \cite{li2021multisports} & Ours      \\ \midrule
Anno type      & Tube       & Tube               & Frame      & Tube   & Tube   & Frame     \\
Classes        & 21         & 24                 & 80         & 10    & 66     & 15        \\
Scene          & Daily life \& Sports & Daily life \& Sports & Daily life & Classroom & Sports & Classroom \\
Avg resolution     & 320 × 240    & 320 × 240       & 320 × 400    & 1450 × 870 & 1280 × 720  & 1600 × 900  \\
Total videos   & 928        & 3194              & 430        & 3343   & 3200     & 4324      \\
Video length   & 1.4sec     & 5.8sec             & 15min      & 17.2sec  & 20.9sec  & 3sec      \\
Total instance & 928        & 4458               & 1.62M      & \textasciitilde 47,320 
& 37701 & 387,860   \\
Multi-Person    & \ding{55}  & \ding{55}    &$\checkmark$    &$\checkmark$ &$\checkmark$ &$\checkmark$   \\
Multi-Label    & \ding{55}    &\ding{55}   &$\checkmark$   &\ding{55} &\ding{55} & $\checkmark$         \\  \bottomrule  \bottomrule
\end{tabular}
\label{tb_stastics}
\end{table*}

\begin{figure*}[!t]
\centering
\centerline{\includegraphics[width=18cm]{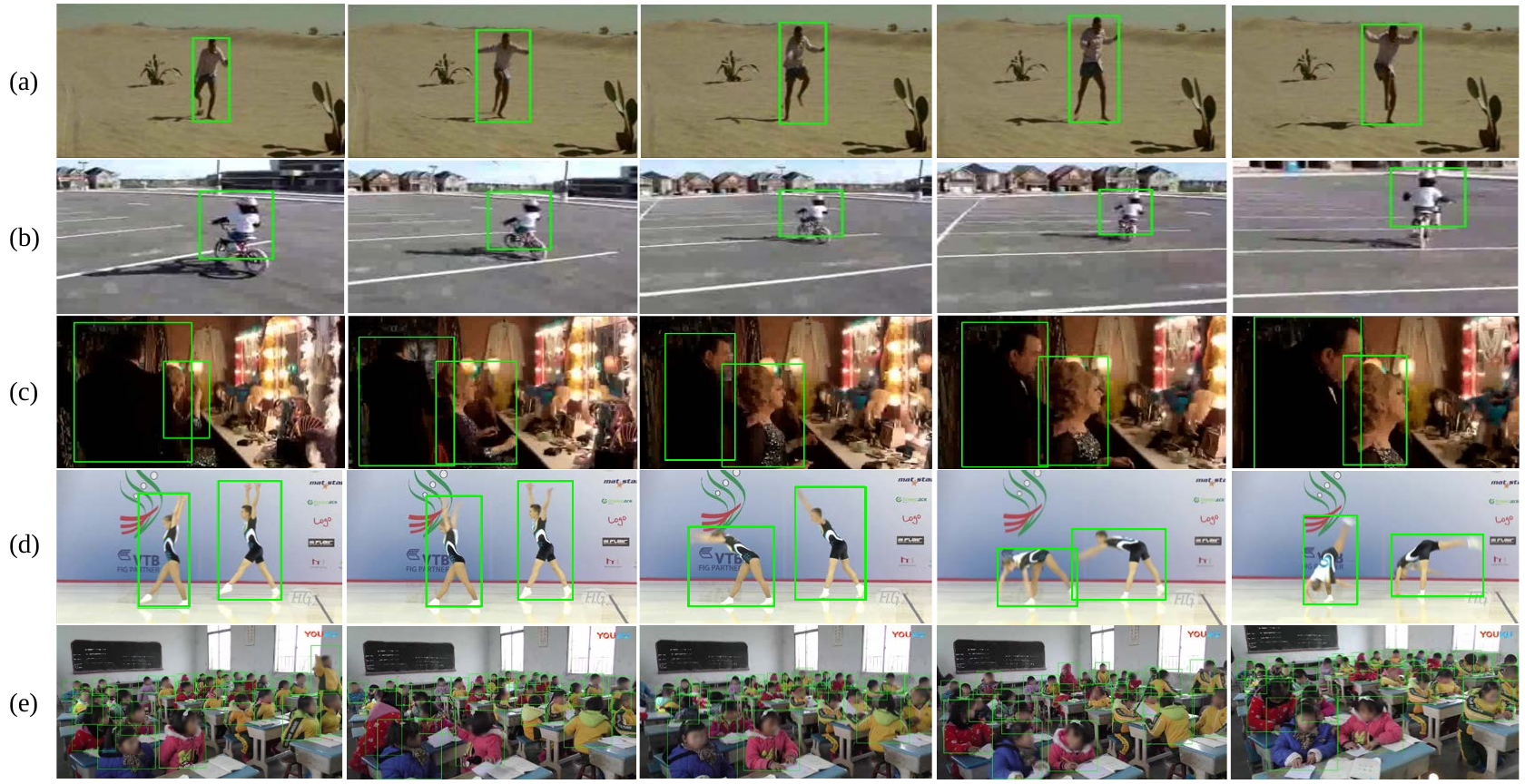}}
\caption{Each row shows samples from J-HMDB \cite{jhuang2013towards}, UCF101-24 \cite{soomro2012ucf101}, AVA \cite{gu2018ava}, MultiSports \cite{li2021multisports} and SAV, respectively. (a) J-HMDB: each video contains one person for a single label. (b) UCF101-24: same as above. (c) AVA: contains multiple persons with multiple labels. (d) MutiSports: contains multiple persons, each with a sports label. (e) SAV: contains dense persons with multiple labels. (Note that: The data from Sun et al. \cite{sun2021student} are not yet publicly available.)}
\label{dataset_compare}
\end{figure*}


Table \ref{tb_stastics} compares the statistics of SAV with the existing action detection datasets.
Large-scale datasets, such as AVA \cite{gu2018ava} and SAV, annotate each clip with a keyframe, and only these two datasets are multi-person and multi-label datasets. In datasets with daily life or sports scenes, AVA \cite{gu2018ava} and Multisports \cite{li2021multisports} provide more detailed categories and instances. The data scale of J-HMDB \cite{jhuang2013towards} and UCF101-24 \cite{jhuang2013towards} is relatively small. Compared with the dataset of Sun et al. \cite{sun2021student} for classroom scenes, SAV has a higher average resolution (1600×900 vs. 1450×870), more fine-grained categories (15 vs. 10), and richer instances (387860 vs. 47320), which poses a new challenge of modeling fine-grained actions with multiple actors and labels in classroom scenarios.



Fig. \ref{dataset_compare} shows sample frames from the comparison dataset and SAV, highlighting that the objects in SAV are significantly smaller and denser than those in the other datasets. Both J-HMDB and UCF101-24 have a single object only. The AVA dataset is annotated at one-second intervals per frame, which captures significant variations in actor movement across consecutive frames. The sports actions in Multisports have large deformations and displacements. SAV shows keyframes of consecutive clips. In contrast, SAV has small displacements in the human object but significant deformations in the detailed motion.

\begin{figure*}[!ht]
\centering
\centerline{\includegraphics[width=17cm]{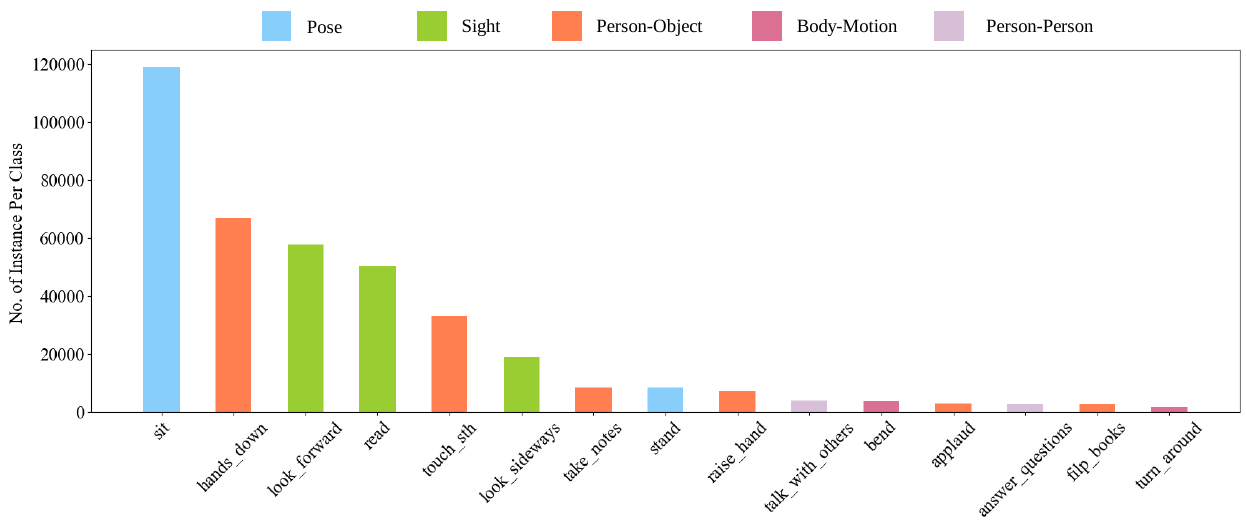}}
\caption{Statistics of each class in the SAV dataset, which is sorted by descending order. Blue for pose actions, green for sight actions, orange for person-object interactions, purple for body-motion actions, and light purple for person-person interactions.}
\label{num_class}
\end{figure*}

Fig. \ref{num_class} illustrates the quantitative properties of our dataset on each category, with the distribution roughly following Zipf's law. As depicted, as a must-have postural action, ``sitting" is much more common than ``standing" in the classroom scenario. Sight actions such as ``reading" and ``looking forward" occur with comparable frequency, while ``looking around" is less common. Among person-object interactions, ``touching" is most frequent, whereas actions like ``flipping a book" are rarer. Additionally, body motion and person-person interactions, such as ``turning around" and ``answering questions", are relatively few compared to the other major categories.

\begin{figure*}[!ht]
\centering
\centerline{\includegraphics[width=18cm]{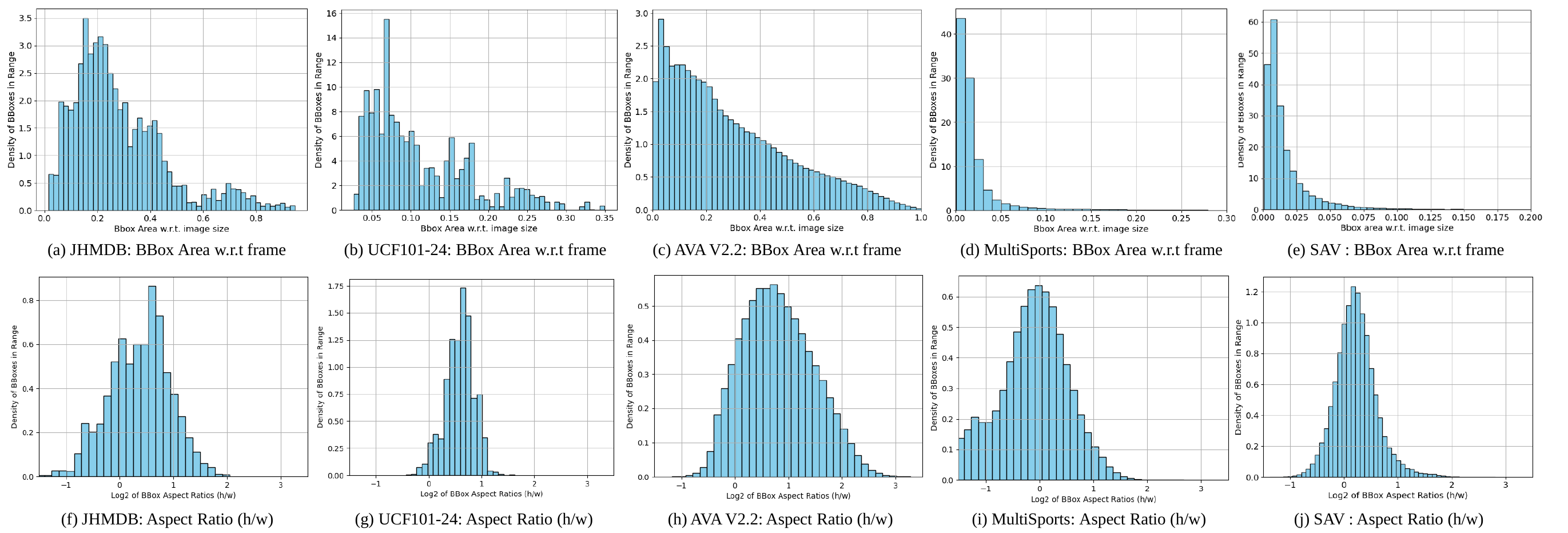}}
\caption{Comparison of the characteristics of bounding boxes. First row: The X-axis implies the ratio of bounding box area w.r.t. video frame. The Y-axis implies the normalized density of bounding box occurrences. Second row: The X-axis implies the aspect ratio of the bounding box area (height/width). The Y-axis implies the normalized density of the bounding box occurrences.}
\label{bbox}
\end{figure*}

Fig. \ref{bbox} shows a comparative analysis of the bounding box properties for different datasets. First, in the first row, the distribution of the proportion of bounding boxes in the J-HMDB and UCF101-24 is mainly between 5\% and 50\% of the image frames, while in AVA, the proportion is mostly below 80\%. In contrast, the bounding boxes in MultiSports and SAV occupy only a very small portion of the image frame, typically less than 1\%. This can make it more difficult to accurately detect and identify motion, as subtle details and fine movements may not be captured effectively. Furthermore, regarding the aspect ratio of the bounding box in the second row, the bounding boxes in J-HMDB, UCF101-24, and AVA typically exhibit a greater height than width. In MultiSports, a large amount of motion deformation results in a wider range of aspect ratios of its bounding boxes. Since the lower part of the human body is usually occluded in classroom scenes, the bounding boxes in SAV often exhibit a slightly larger height relative to the width. 
Although the bounding boxes of SAV have less variation in aspect ratio, their density is much larger than other datasets, posing a challenge to the completeness of detection.

\subsection{Data characteristics}
\begin{figure}[!t]
\centering
\centerline{\includegraphics[width=9cm]{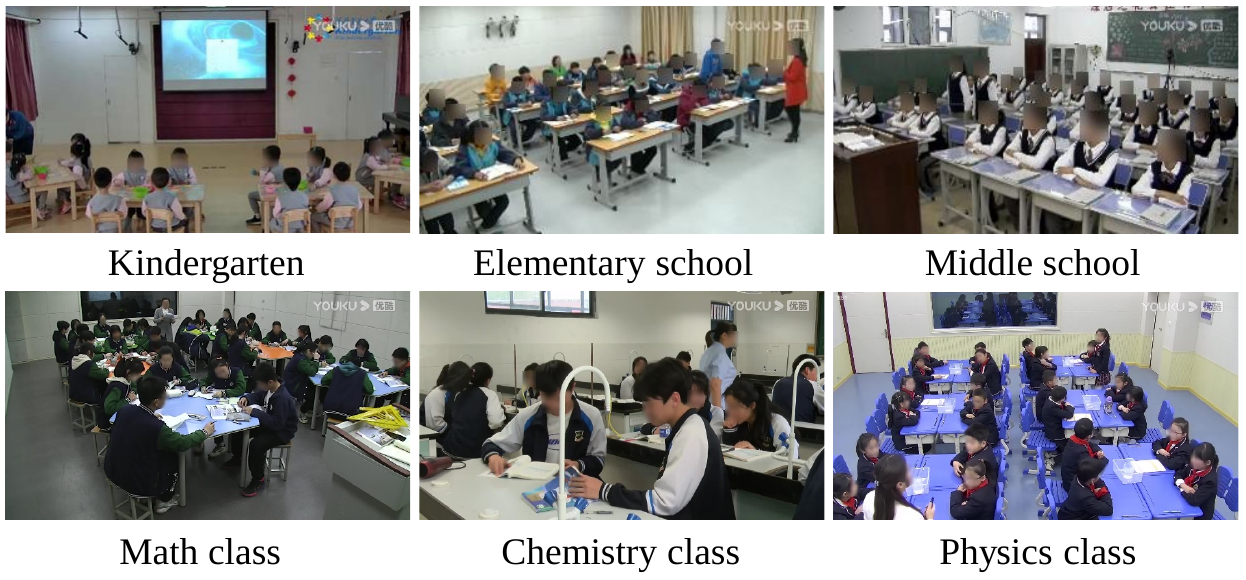}}
\caption{First row: the different educational stages of classrooms in SAV: kindergarten, elementary school, and middle school. Second row: the different course scenarios in SAV, such as math, chemistry, and physics.}
\label{different_stage}
\end{figure}

\textbf{Diverse and real classroom scenes.}
The dataset contains a wide range of real classroom scenes, fully capturing the actions that occur in natural teaching environments. As shown in Fig. \ref{different_stage}, the dataset contains a range of classroom settings from kindergarten to middle school, covering critical periods of student development. In addition, it also includes scenes from different teaching courses such as math, chemistry, physics, Chinese, English, computer science, and fine arts. This diversity provides rich scenarios that can significantly advance algorithm research.

\textbf{High Quality.}
The videos of SAV are in high resolution (720P or 1080P), preserving the details of small persons and objects. In addition, our annotations are accurately labeled and cross-checked by professionals with advanced degrees. Professional annotators and careful quality control ensure that the annotations are both consistent and highly accurate.

\begin{figure}[!ht]
\centering
\centerline{\includegraphics[width=5cm]{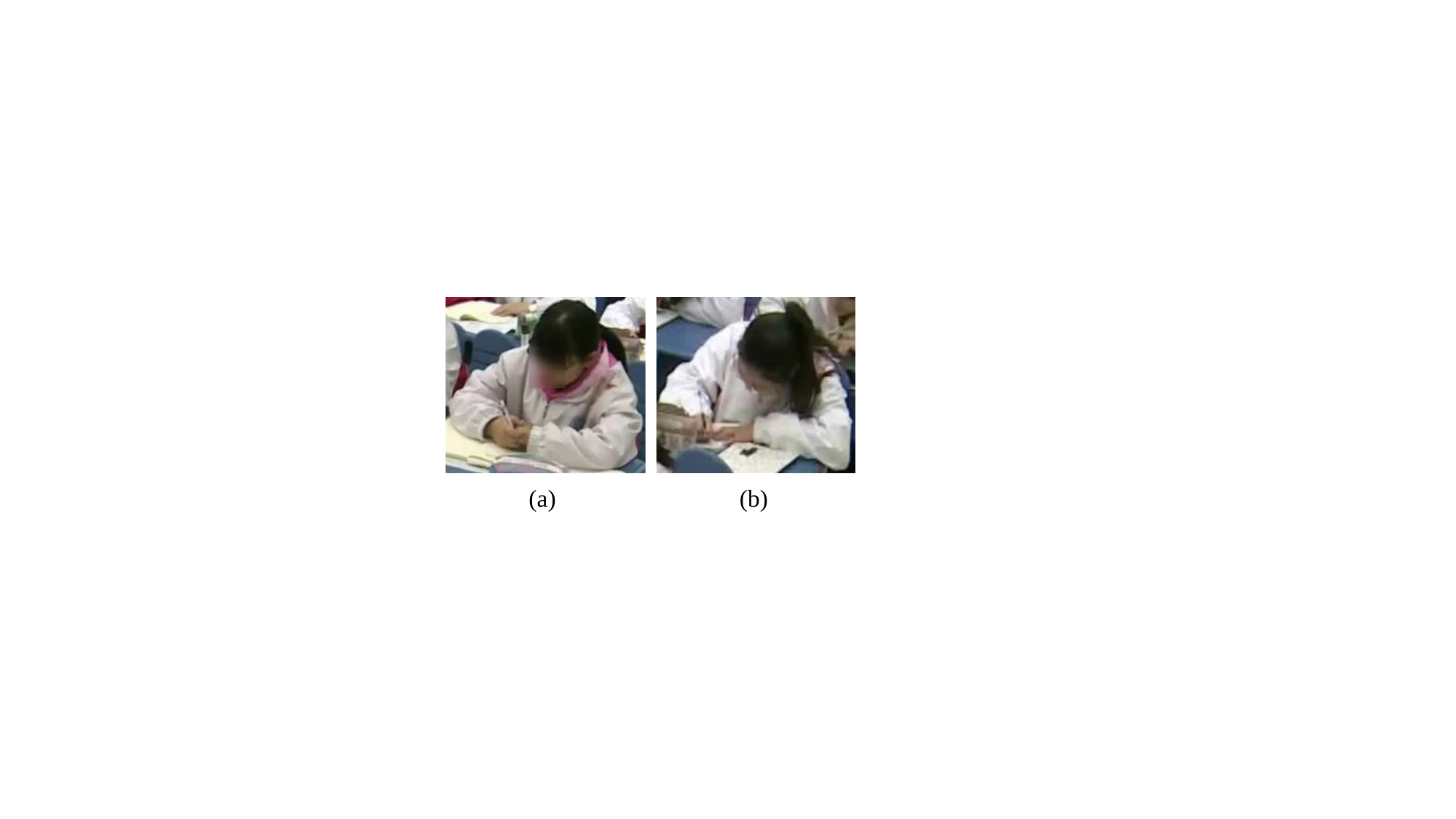}}
\caption{The similarity between different action categories in SAV: (a) touch something, (b) take notes.}
\label{class_similarity}
\end{figure}

\textbf{Challenging.}
As discussed above, SAV is challenging in several aspects compared to existing datasets: 

(1) Multi-label.
Student actions in the classroom are categorized into multiple fine-grained labels. A single subject may be associated with several labels, which requires the algorithm to accurately focus on different regions of interest in the video, such as the hands and eyes. In addition, different actions often exhibit visual similarities. Fig. \ref{class_similarity} illustrates a comparison between two different actions, where the person-object interaction in Fig. \ref{class_similarity} (a) is touching something (pen), and the person-object interaction in Fig. \ref{class_similarity} (b) is taking notes. This similarity requires the algorithm to have strong feature extraction and learning capabilities to distinguish similar but different actions more accurately.

(2) Multi-object.
The typical classroom scenario often presents multiple objects, as shown in the last row of Fig. \ref{dataset_compare}. The classroom scenes in the SAV dataset have up to 68 participants. This high density of objects challenges the completeness and accuracy of detection tasks. Moreover, this multi-person scenario with different concurrent actions prevents the model from distinguishing action classes only by background, which requires the model to capture subtle action changes.


\begin{figure}[!ht]
\centering
\centerline{\includegraphics[width=6cm]{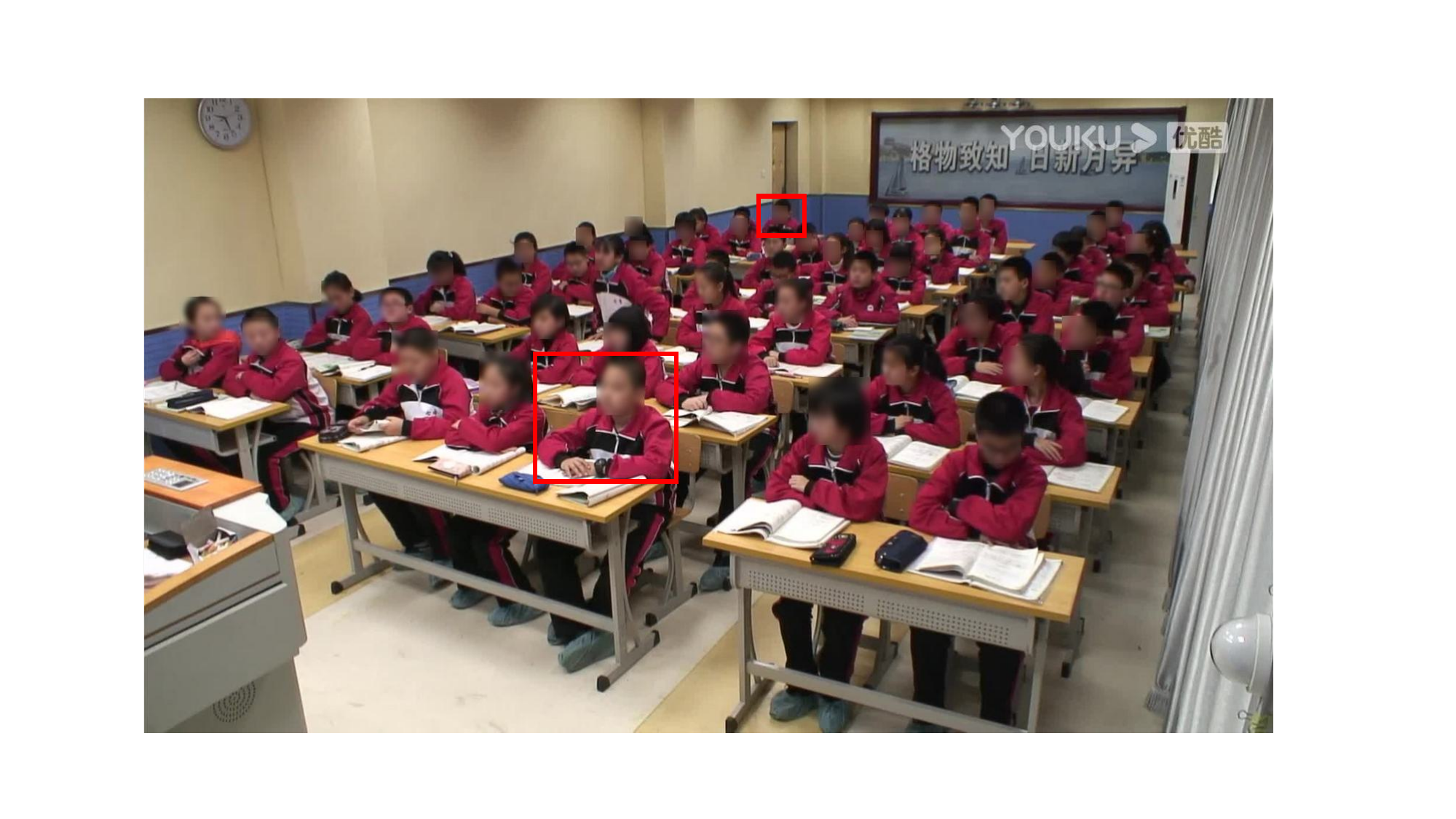}}
\caption{The illustration of large scale differences in SAV, where the red boxes are used as a mark to highlight the difference.}
\label{scale_difference}
\end{figure}

(3) Large scale differences.
Due to the fixed position of surveillance cameras, there is a significant variation in the size of students seated in the front and back rows of the classroom. As illustrated in Fig. \ref{scale_difference}, this scale difference poses a challenge for the detection and analysis of student actions. Existing methods often struggle to address this problem, as they tend to focus primarily on the larger student objects in the front rows while neglecting the smaller objects in the back rows. Therefore, algorithms that can adapt to student objects of different scales are crucial for accurately analyzing student actions.

\begin{figure}[!ht]
\centering
\centerline{\includegraphics[width=9cm]{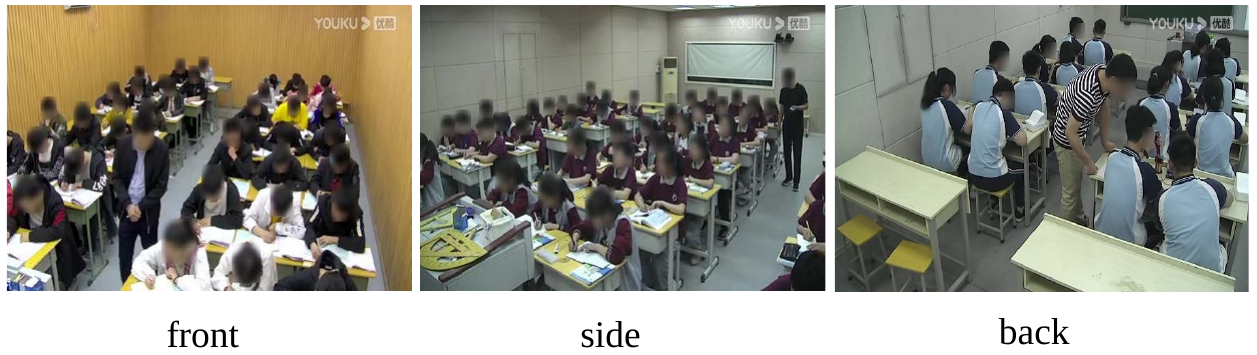}}
\caption{The illustration of different shooting angles in SAV, including front, side, and back.}
\label{shooting_angles}
\end{figure}

\begin{figure}[!ht]
\centering
\centerline{\includegraphics[width=9cm]{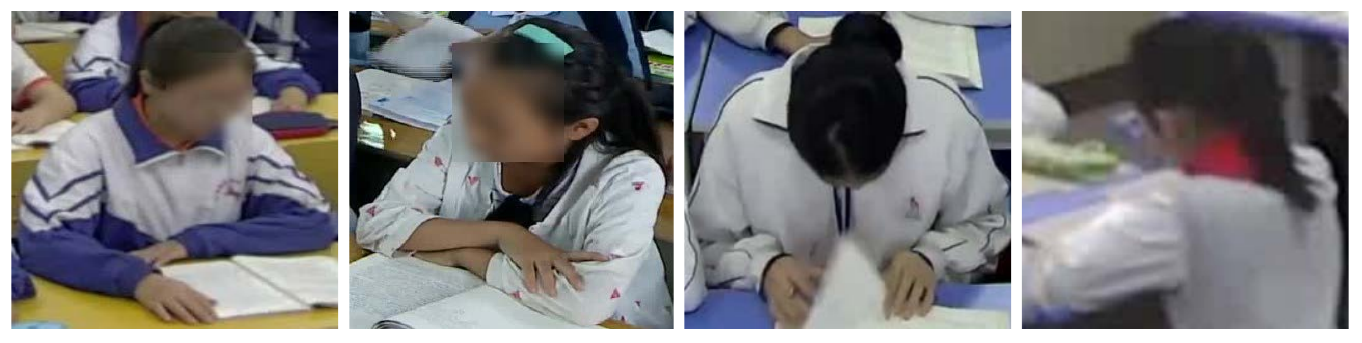}}
\caption{The illustration of differences in perspectives presented by the same category of “reading” in different classroom settings.}
\label{read}
\end{figure}
(4) Varying shooting angles.
In the SAV dataset, video recordings of different classrooms are taken from various angles, mainly including front, side, and back views, as shown in Fig. \ref{shooting_angles}. Therefore, the same action may exhibit completely different visual characteristics. Fig. \ref{read} demonstrates the variation in the category of ``reading" across different classroom settings. Such diversity requires action analysis algorithms to be robust and adaptable to classify actions accurately under varying observational conditions.

\begin{figure}[!ht]
\centering
\centerline{\includegraphics[width=9cm]{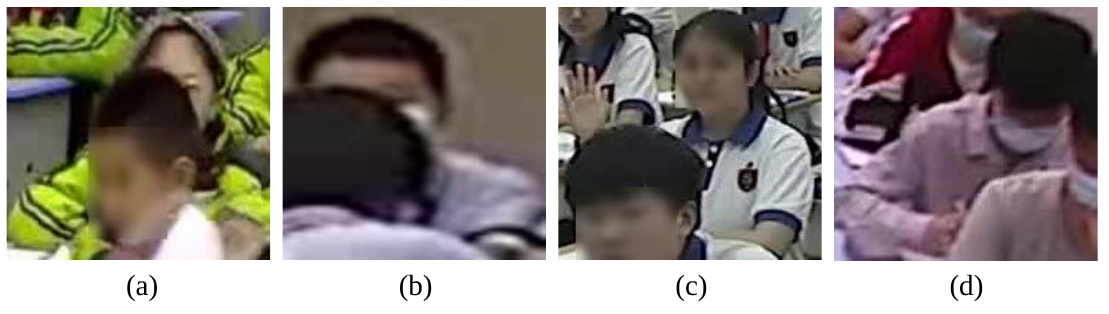}}
\caption{The illustration of occlusion in SAV. The person in (a) and (b) is occluded by the front row while looking forward, the hand of the person in (c) is occluded while raising her hand, and the hand of the person in (d) is occluded while taking notes. }
\label{occlusion}
\end{figure}

(5) Occlusion.
In a classroom environment, the hands and faces of students may be occluded by objects such as desks and students in the front row, as shown in Fig. \ref{occlusion}. This heavily occluded situation poses a challenge for algorithms to recognize student actions, which requires algorithms to be able to process partially visible action information and enhance their robustness against such occlusions.


\section{Action detection model}

\begin{figure}[!ht]
\centerline{\includegraphics[width=9cm]{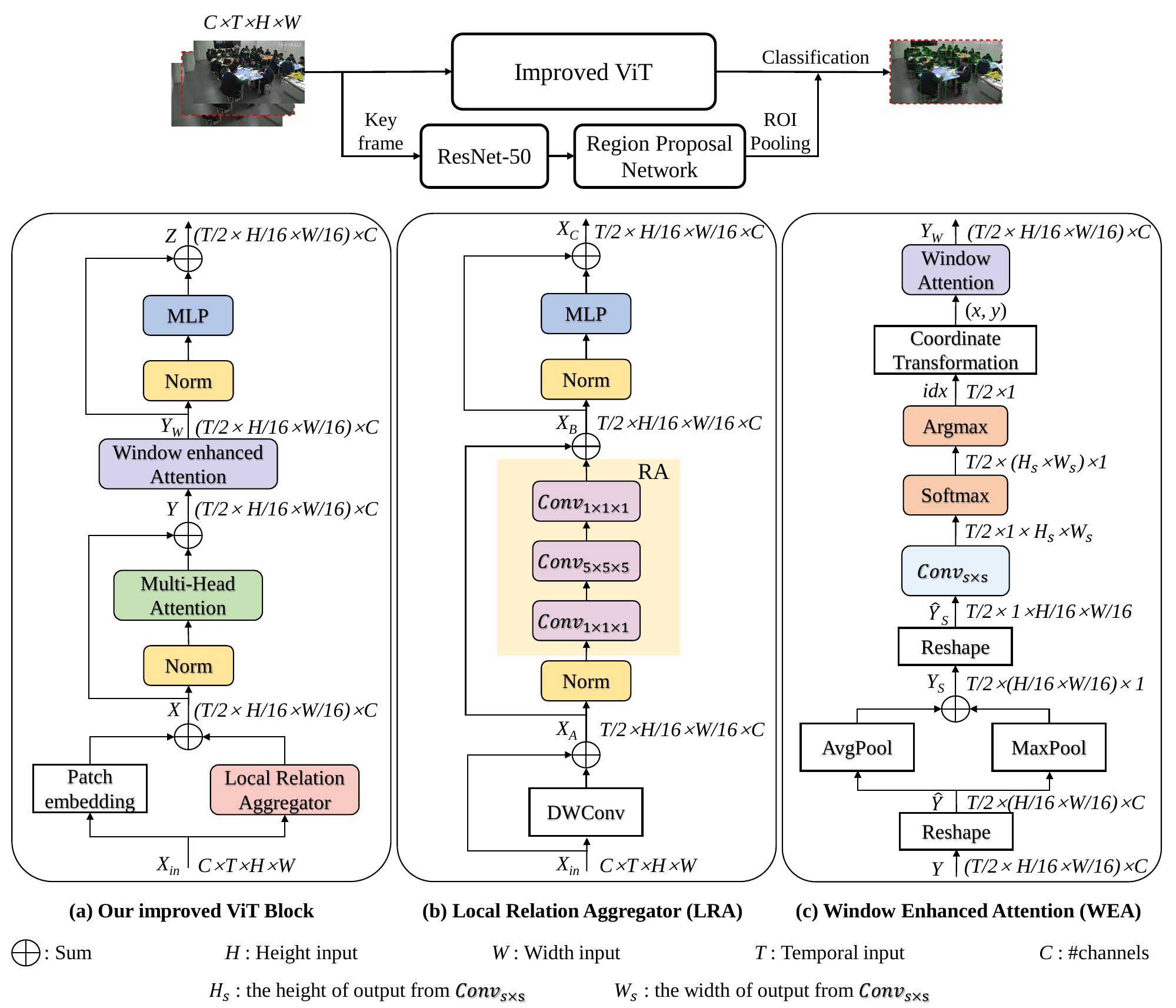}}
\caption{The framework of our method for action detection. (a) Our improved ViT block, (b) local relation aggregator (LRA), (c) window enhanced attention (WEA).}
\label{framework}
\end{figure}

ViT leverages a self-attention mechanism to capture global context and long-range dependencies in images. This mechanism allows ViT to excel in scenarios where the main object is large, distinct, and set against a non-complex background, enabling the model to focus effectively on prominent elements within the image. However, the attention mechanism of ViT is uniformly distributed over the whole image. It does not achieve significant focusing and filtering power over local features with convolutional kernels as CNN does. 
As a result, when faced with scenes with small and dense objects like SAV, the global focus of ViT may make it difficult to capture important local dynamics and tiny objects. 

Inspired by recent research on unifying convolution and self-attention mechanisms \cite{li2023uniformer}, we propose an improved ViT network for action detection that effectively fuses global and local visual information. As shown in Fig. \ref{framework}, we first feed the T-frame image into the improved ViT to extract discriminative features, and feed the keyframe in the T-frame into Faster-RCNN \cite{ren2016faster} with ResNet-50 as the backbone. Subsequently, the Region Proposal Network (RPN) generates a series of region proposals containing objects on the feature map. We then fix and map these region proposals to the feature map output by the improved ViT module through RoI Pooling, and expand the 2D RoI to 3D RoI at the same spatial position for the final classification decision.
As shown in Fig. \ref{framework} (a), our improved ViT block contains four key modules: Local Relation Aggregator (LRA), Multi Head Attention module (MHA), Window Enhanced Attention (WEA), and Multilayer Perceptron (MLP):
\begin{equation}
\label{X}
    X = PE(X_{in}) + LRA(X_{in}),
\end{equation}
\begin{equation}
\label{Y}
    Y = MHA(Norm(X)) + X,
\end{equation}
\begin{equation}
\label{Z}
    Z = MLP(Norm(WEA(Y))) + WEA(Y).
\end{equation}

First, the input T-frame image $X_{in}$ is divided by patch embedding (PE) into patches of the size  $(T/2 \times H/16 \times W/16) \times C$. Here, $H$ and $W$ denote the height and width of the input video frame, respectively. $T$ represents the number of sampled frames, and $C$ is the channel dimension. Concurrently, the image is processed by a local relation aggregator module, which integrates a depthwise Convolution (DWconv), a relation aggregator (RA), and an MLP. DWconv dynamically integrates the location information into all tokens, while RA enhances local feature representation by learning local token affinities at a shallow level, which is particularly beneficial in dense object scenes. The LRA can help the model understand the image content at a small scale. 

Following these initial steps, the tokens derived from PE are combined with those from LRA to supplement them with local details within the token and neighborhood relationships between tokens (Eq. (\ref{X})). These combined tokens are then subjected to the global MHA layer for deep-level token affine learning (Eq. (\ref{Y})). To further enhance the detail of tiny objects, the Window Enhanced Attention (WEA) module is employed. This module improves the capture of localized video features by identifying and focusing on the most responsive window regions within a token, and applying additional spatio-temporal attention to these regions. This approach naturally augments local information in the global without the need for redundant fusion of global-local features. Subsequently, an MLP layer is added as in conventional ViT, which consists of two linear layers and a nonlinear function GeLU (Eq. (\ref{Z})). 



\subsection{Local Relation Aggregator}
We employ a local relation aggregator module fused with the features after patch embedding in the vanilla ViT to enhance the local dynamic perception of small and dense objects in SAV. As shown in Fig. \ref{framework} (b), given the input video $X_{in}\in\mathbb{R} ^ {C \times T \times H \times W}$, the LRA block uses DWConv with a convolutional kernel size of $3 \times 3 \times 3$ for dynamic position embedding. After normalization, the features are transformed through a convolutional layer with a convolutional kernel of $1 \times 1 \times 1$, which is used to recombine the channel features while maintaining the feature dimensions. 
To learn the local token affinity within a small neighborhood, which is similar to the design of a convolutional filter, we use a convolutional layer with a kernel size of $5 \times 5 \times 5$ to aggregate local relationships. This local relationship depends only on the relative positions between tokens. In addition, this convolutional layer processes each channel independently by grouping the convolutions to reduce the number of parameters. The feature is then concatenated with the original feature for multilayer perception. This multilayer perceptron consists of two $1 \times 1 \times 1$ convolutional layers and a GELU activation function to further process and optimize the features. The process can be expressed as
\begin{equation}
\label{Local Relation Aggregator}
    X_{A} = DWConv(X_{in}) + X_{in},
\end{equation}
\begin{equation}
\label{Local Relation Aggregator}
    X_{B} = RA(Norm(X_{A})) + X_{A},
\end{equation}
\begin{equation}
\label{Local Relation Aggregator}
    X_{C} = MLP(Norm(X_{B})) + X_{B},
\end{equation}
where RA denotes the relation aggregation operation consisting of two convolutional layers with kernel size $1 \times 1 \times 1$ and one convolutional layer with kernel size $5 \times 5 \times 5$.

\subsection{Window Enhanced Attention}
Vanilla ViT evaluates the interdependence between tokens through the self-attention mechanism.  However, in complex SAV scenes, discriminative tiny objects often occupy only a small area within a token.  ViT tends to overlook these minute details. Therefore, we propose to focus on the most informative regions in each token and strengthen the representation learning of these regions.
The structure of the Window Enhanced Attention module is shown in Fig. \ref{framework} (c). To achieve the requirement of locating the window with the largest local response, the feature $Y \in \mathbb{R} ^ {(T/2 \times H/16 \times W/16) \times C}$ is first reshaped to $\hat{Y} \in \mathbb{R} ^ {T/2 \times (H/16 \times W/16) \times C}$ for subsequent spatial manipulation, and then global average pooling and global maximum pooling are performed on the channel dimension to obtain two pooled features. These two features are summed to merge the channel information and generate a preliminary spatial response map $Y_{S}$:


\begin{equation}
\label{Sum}
    Y_{S} = Sum( AvgPool(\hat{Y}), MaxPool(\hat{Y})),
\end{equation}
where $Y_{S}\in \mathbb{R} ^ {T/2 \times (H/16 \times W/16) \times 1} $. $AvgPool(\cdot)$ and $MaxPool(\cdot)$ represent global average pooling and global maximum pooling based on channels. $Y_{S}$ is then reshaped to $\hat{Y_{S}} \in \mathbb{R} ^ {T/2 \times 1 \times H/16 \times W/16}$ for subsequent spatial convolution operation.

The response map $\hat{Y_{S}}$ is passed through an unfilled convolutional layer for local feature extraction to capture the salient features associated with the window. The convolution kernel size $s$ is the same as the window size $w$. Subsequently, we normalize these feature responses using the Softmax function to identify the region with the maximum response. This maximum response location is calculated by coordinate transformation to obtain the coordinates of the upper left corner of the region of interest for each video frame. These coordinates indicate the exact location in the video where the critical action occurs. The index value with the largest response can be achieved by the Argmax operation:

\begin{equation}
\label{Softmax}
    idx = Argmax(Softmax( Conv_{s\times s}(\hat{Y_{S}}))),
\end{equation}
where $idx$ indicates the corresponding index values for the maximum response of each image frame. $Argmax(\cdot)$ is the function that maximizes the argument. $Conv_{s\times s}$ is an unfilled convolution kernel of size $s$.


In the WEA, the coordinate transformation converts the one-dimensional maximum response position index to two-dimensional image coordinates $(x, y)$. It maps the maximum response position index obtained from the convolution operation back to the exact position of the original image frame to determine the top-left coordinates of the region of interest. The process of coordinate transformation can be expressed as
\begin{equation}
\label{Softmax}
    (x,y) = (mod(idx, u), \lfloor(idx,u)\rfloor),
\end{equation}
where $(x,y)$ denotes the horizontal and vertical coordinates of the upper-left corner of the original image frame with the maximum response region. $u= H/16 - s + 1$ is the size of the feature map after the convolution layer. $mod(\cdot)$ is the modulo operation of the $idx$ divided by $u$. $\lfloor\cdot\rfloor$ indicates the integer division representation between the $idx$ and $u$.

The obtained attention region coordinates $(x,y)$ are used to cut out the corresponding windows from each token that contain the most important visual information in the video. The features of these specific windows are then further enhanced by a spatio-temporal attention mechanism to highlight the critical spatio-temporal information. This process can be expressed as
\begin{equation}
\label{Softmax}
    Y_{W} = attn(Y(x:x+w, y:y+w)),
\end{equation}
where $Y_{W}$ is the feature after the window enhanced attention module. $attn$ is the space-time self-attention mechanism. 

Together, the LRA and WEA form a pipeline in which the LRA strengthens the overall local detail representation at a shallow level, while WEA refines and focuses on the most relevant regions of these local details at a deep level, ensuring that local feature representations are adequately captured. By emphasizing local feature representation, these modules aim to collectively improve the ability of the model to handle subtle dynamics and small-scale objects in dense and complex scenes.

\section{experiments}
\subsection{Datasets and Metrics}
\textbf{SAV benchmark.} To construct a reliable action detection benchmark, we manually select instances from 758 publicly released classroom videos and split these instances into training and test sets. The current benchmark covers 308,962 training instances out of 3459 clips, 78,898 test instances out of 865 clips, and 15 categories. The allocation ratio is close to 4:1. video resolutions of all instances are mainly standard 720P and 1080P, and also include some irregular video resolutions. In addition, the bounding box annotations we provide are all expressed in relative coordinates.

\textbf{Metrics.} Since the videos in SAV are trimmed, we use frame-mAP to evaluate action detection performance. The Intersection-over-Union (IoU) threshold is set to 0.5 for frame-mAP, which represents the ratio of the intersection area to the concatenation area of the real ground bounding box and the corresponding detected bounding box \cite{ren2016faster}. For each class, we report the mean Average Precision (mAP) which is calculated by averaging over all classes.\\

\subsection{Action Detection Results}

We evaluate several representative action detection methods on the SAV and AVA datasets, and the results are detailed in Table \ref{result_SAV} and Table \ref{result_AVA}, respectively. For the YOWO v2 \cite{yang2023yowov2}, we employ the officially released code and the 3D-ShuffleNetv2-1.0x backbone, setting the batch size to 8. The Slowfast, MViT, and MViTv2 are run using the Pyslowfast framework \cite{fan2020pyslowfast}, and the VideoMAE is run using the MMAction2 framework \cite{2020mmaction2}. Their batch sizes are all set to 8 and VideoMAE uses the vit-base configuration.
Our method incorporates 5 LRA blocks with 320 channels, 8 LRA blocks with 768 channels, and 12 ViT blocks containing WEA. We use pre-training weights from the VideoMAE with a batch size of 2 and subsequently fine-tune them for our specific application. The pre-trained weights for all methods are based on Kinetics-400. To verify the effectiveness of self-supervised learning, the pre-trained weights of the VideoMAE are not additionally fine-tuned using labels from Kinetics-400.
In addition, in the spatial domain, we follow the setting of \cite{fan2020pyslowfast} to scale the short side of the video to 256 pixels and perform a $224 \times 224$ center crop. In the temporal domain, we uniformly sample a clip from the full-length video, and the input of the network consists of T frames with a time step of $\tau$, denoted as $T \times \tau$.
The window size for our method is set to 7. Owing to the multi-label nature of the dataset, we substitute the standard softmax loss function with a sum of binary sigmoid losses, one for each class.


As shown in Table \ref{result_SAV} and \ref{result_AVA}, our method achieves the highest mAP of 67.90\% on SAV and 27.4\% on AVA. Our method specifically boosts attention to small and dense objects in SAV by capturing local dynamics with the LRA and focusing on discriminative small objects with the WEA module. 
The AVA dataset, in contrast, primarily consists of larger-scale human actions, which are more global and coarse-grained in nature. For example, actions like "running" or "jumping" involve broader spatial and temporal contexts and do not rely as heavily on subtle details or small objects. Therefore, our method provides limited improvements on the AVA dataset.
In addition, due to the dense objects in SAV, we need to use deep LRA to capture the local dynamic information of multiple objects, which leads to higher FLOPs and parameters. 
The YOWOv2 shows suboptimal performance on both SAV and AVA datasets. YOWOv2 uses a lightweight backbone such as 3D ShuffleNetv2 to capture spatio-temporal features, which improves the speed, but the design of its lightweight 3D convolution model has limitations in extracting motion information features. The fusion of spatial and temporal features in YOWOv2 is also not sufficient.

The SlowFast model effectively captures spatial and temporal information by leveraging dual pathways. Utilizing ResNet 50 and ResNet 101 as backbones, the SlowFast significantly outperforms the YOWOv2, illustrating that its dual-path design yields a more comprehensive understanding of video content. However, this design focuses mainly on the temporal dynamics of individual actions or simple scenes and lacks mechanisms to deal with small objects and local details. It is not suitable for complex action combinations and intensive interactions in multi-label classroom scenarios.
The MViT and MViTv2 enhance the vanilla ViT with a multi-scale mechanism, showing promising performance. The MViTv2 improves upon MViT by optimizing position embedding and residual pooling connections. However, the scaling down of feature maps in their hierarchical design often leads to a loss of spatial resolution, making it challenging to distinguish small and densely packed objects.
The VideoMAE achieves satisfactory results by learning to reconstruct the masked parts of the videos. Without fine-tuning with Kinetics-400 labels, it achieves 66.56\% and 26.7\% mAP on the SAV and AVA datasets, respectively. While it has the advantage of capturing global video context, the reliance on a vanilla ViT backbone makes it difficult to efficiently learn multiple small-scale local details in dense scenes.

\begin{table}[]
\caption{Frame-mAP of the state-of-the-art comparison methods and our methods on SAV. The $\checkmark$ in Ex. Labels indicates supervised data is used for pre-training while \ding{55} indicates only unlabeled data is used for the pre-training. $T \times\tau$ refers to the frame number and corresponding sample rate.}
\label{result_SAV}
\centering
\begin{tabular}{cccccc}
\toprule
Method        &Ex. Labels  & $T \times\tau$  & mAP   & FLOPs & Params \\ \midrule
YOWOv2        &\checkmark & 24×3 & 24.99 & 13G   & 52M                        \\
SlowFast R50  &\checkmark & 32×3 & 34.97 & 50G   & 33M                        \\
MViT\_B\_24   &\checkmark & 16×4 & 39.27 & 98G  & 52.9M                        \\
MViT\_B\_24   &\checkmark & 32×3 & 41.45 & 236G  & 52.9M                        \\
MViTv2\_B     &\checkmark & 32×3 & 45.11 & 225G  & 51M                        \\
VideoMAE      &\ding{55} & 16×4 & 66.56 & 180G  & 87M                        \\ \midrule
Ours          &\ding{55} & 16×4 & \textbf{67.90} & 412G  & 172M                       \\\bottomrule
\end{tabular}
\end{table}

\begin{table}[]
\caption{Frame-mAP of the state-of-the-art comparison methods and our methods on AVA v2.2. The $\checkmark$ in Ex. Labels indicates supervised data is used for pre-training while \ding{55} indicates only unlabeled data is used for the pre-training.}
\label{result_AVA}
\centering
\begin{tabular}{@{}cccccc@{}}
\toprule
Method        &Ex. Labels & $T \times\tau$    & mAP  & FLOPs & Params \\ \midrule
YOWOv2        &\checkmark & 32×3 & 18.7 & 13.7G   & 52M    \\
SlowFast R101 &\checkmark & 8×8  & 23.8 & 138G  & 53M    \\
MViT\_B       &\checkmark & 16×4 & 24.5 & 70.5G  & 36.4M  \\
MViT\_B       &\checkmark & 32×3 & 26.8 & 170G  & 36.4M  \\
MViTv2\_B     &\checkmark & -    & -    & -     & -      \\
VideoMAE      &\ding{55} & 16×4 & 26.7 & 180G  & 87M    \\ \midrule
Ours          &\ding{55} & 16×4 & \textbf{27.4}  & 412G  & 172M   \\ \bottomrule 
\end{tabular}
\end{table}

\subsection{Ablation study}

\begin{figure}[!ht]
\centerline{\includegraphics[width=8cm]{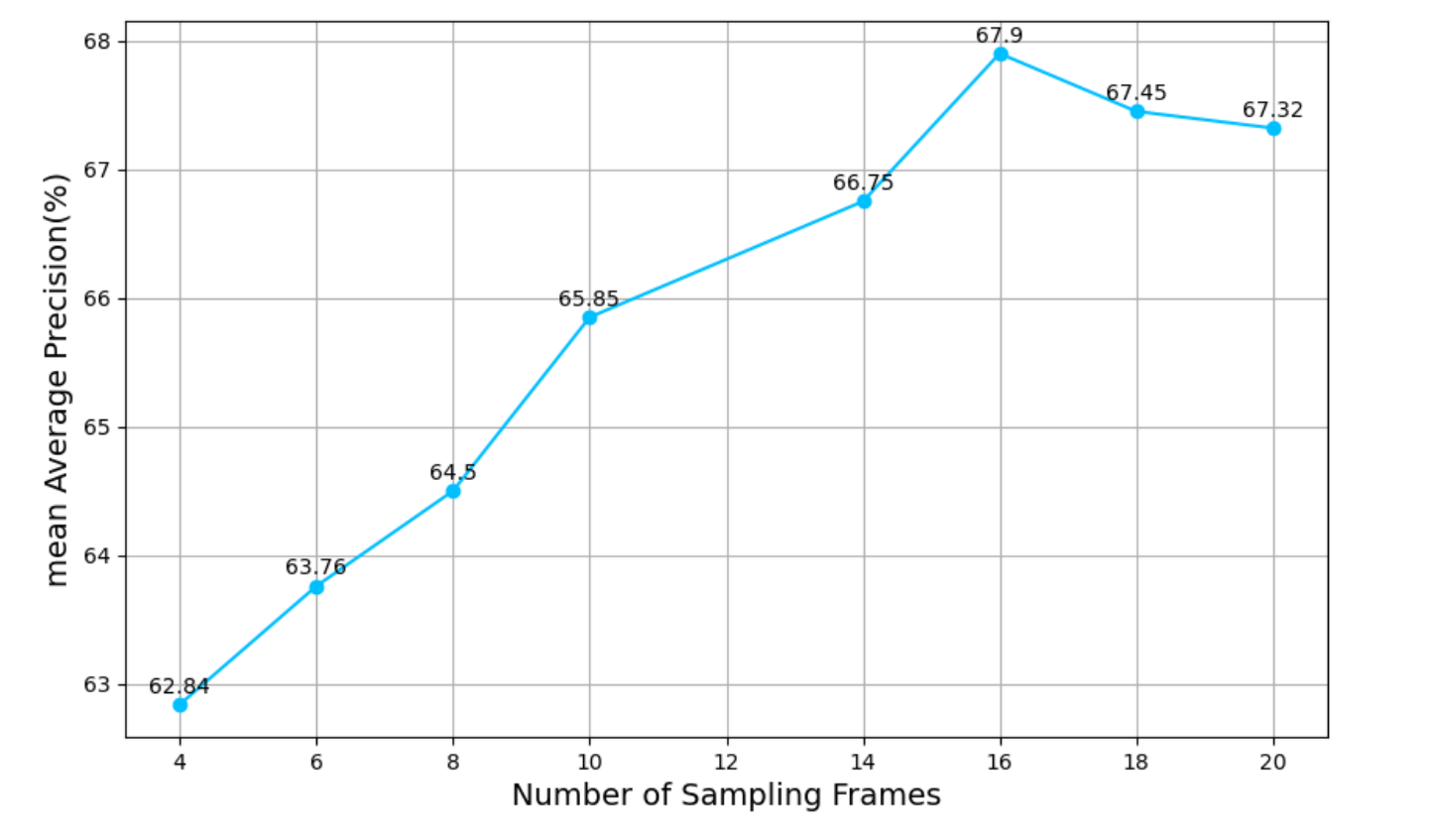}}
\caption{The impact of the number of sampling frames on performance on SAV.}
\label{frame}
\end{figure}

\textbf{How important is temporal information for SAV?}
As shown in Fig. \ref{frame}, the accuracy of our proposed method improves with the increase of the number of sampled frames. This suggests that incorporating more spatio-temporal information from the SAV dataset can enhance the video representation and thus improve the performance. The performance saturates at 16 frames and then decreases with further increase in the number of frames. This decrease occurs because the objects in the class usually remain constant or change slowly. Therefore, too many frames fail to provide additional useful information but instead introduce redundant information, which negatively affects the training effect of the network. This also shows that our manually selected video length of three seconds is sufficient to capture key information of various actions.

\begin{figure}[!ht]
\centerline{\includegraphics[width=9cm]{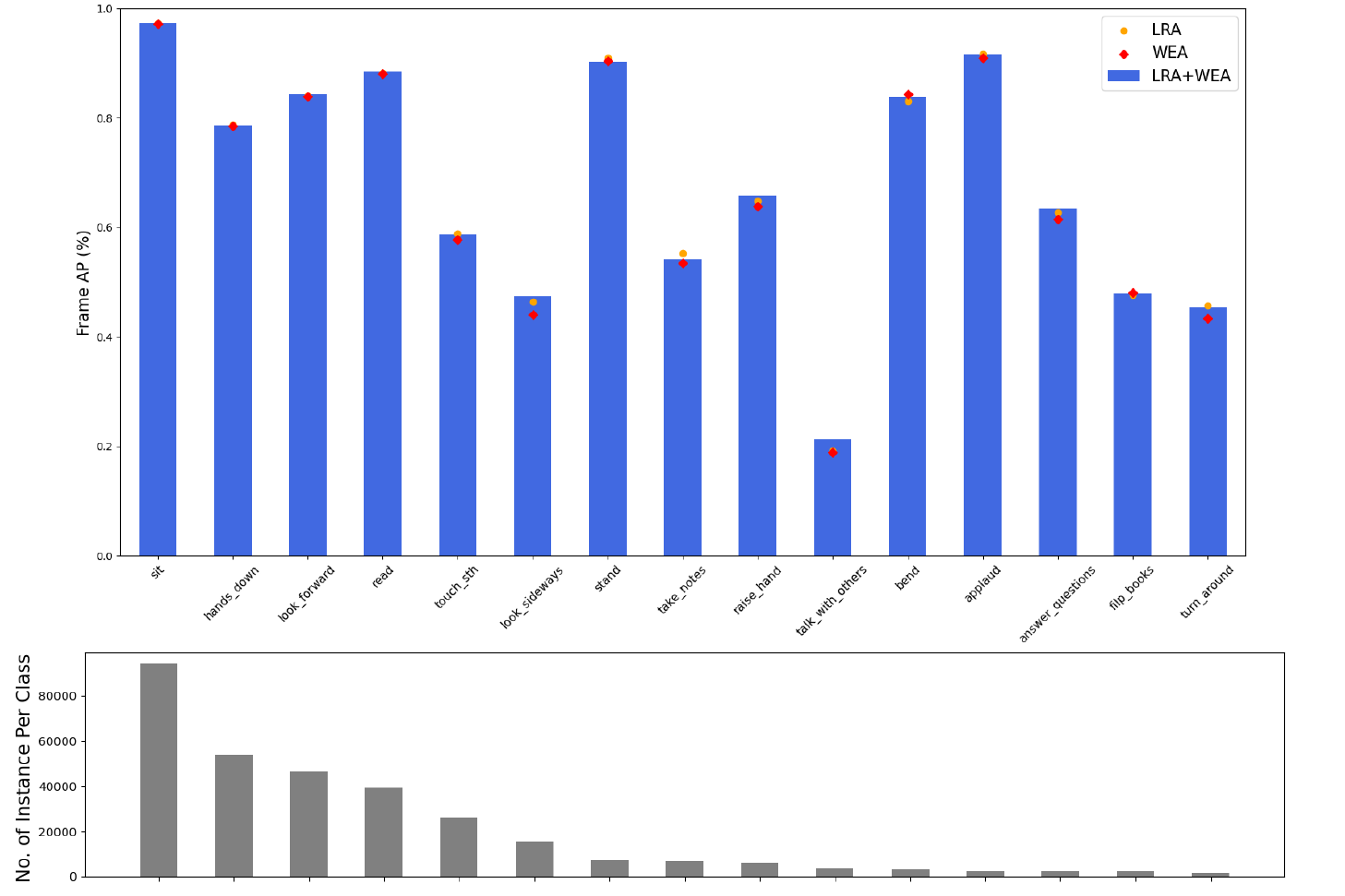}}
\caption{Top: the performance of models for each action class, sorting by the number of training instances. Bottom: the number of training instances per class. Although more training data is better, not all classes have the same complexity. For example, ``applaud", one of the smallest classes, has high performance due to its low diversity.}
\label{class_map}
\end{figure}

\textbf{Which action categories are challenging?}
The top diagram in Fig. \ref{class_map} illustrates the performance of our method on different categories when using the LRA module, the WEA module alone, and the two modules combined. The diagram below shows the number of samples for different categories in the training set. In general, an increase in the number of training samples enhances the performance of the model, but not all categories have the same complexity. For example, the scene-related ``sit” and ``stand” categories show higher performance levels. Categories with lower diversity, such as ``bend” and ``applaud”, also perform well despite having fewer samples. However, some fine-grained categories, such as ``touch\_sth” and ``look\_sideways”,  are still difficult to be accurately distinguished by the model even with a larger training sample. In addition, the high intra-class diversity in the ``talk\_with\_others" category poses additional challenges to the model.

\textbf{How important are different attention styles in the WEA?}
Table \ref{Attention} shows the impact of various attention mechanisms within the WEA module on the performance of the model on the SAV dataset. 
The space-only method produces higher mAP than the time-only method. This suggests that in classroom settings, capturing temporal variations of tiny objects alone is not enough to fully understand the semantics of actions. Spatial information provides key visual cues about object morphology and scene structure, which are essential for parsing the full semantics of actions.
The divided space-time method achieves a mAP of 66.94\%, with a parameter count of 201M and 421G FLOPs. This higher mAP suggests that distinct spatial and temporal attention can enhance detection accuracy, although at the cost of increased model complexity and computational demand. 
Compared to the divided space-time method, the joint space-time method has the highest mAP of 67.90\%, with fewer parameters (172M) and slightly reduced FLOPs (412G). This indicates that integrating spatial and temporal attention effectively improves model performance without increasing computational complexity.
Overall, the joint space-time attention shows the best results on the SAV dataset.

\begin{table}[!ht]
\caption{Different attention schemes in Window enhanced attention on SAV.}
\label{Attention}
\centering
\begin{tabular*}{0.45\textwidth}{@{\extracolsep{\fill}}llll@{}}
\toprule 
Attention          & mAP   & Params & FLOPs \\ \midrule
Space only         & 66.37 &  172M      & 410G      \\
Time only          & 66.21 &  172M      & 409G      \\
Divided Space-Time & 66.94 &  201M      & 421G      \\
Joint Space-Time   & \textbf{67.90} & 172M       & 412G      \\ \bottomrule 
\end{tabular*}
\end{table}

\textbf{How important is the integration of average pooling and maximum pooling in the WEA?}
Table \ref{fusion} describes the impact of different fusion methods involving avgpool and maxpool in the WEA module on the SAV dataset. The results show that feature fusion using the summation operation significantly outperforms the concatenation method in terms of mAP without introducing additional parameters or computational complexity. Maximum pooling preserves salient information but loses detail, and average pooling preserves detailed information but is insensitive to salient information. This complementary relationship allows the summing operation to find the region with the largest feature response more accurately than direct concatenation, which is critical for focusing on discriminative tiny objects in the SAV. 

\begin{table}[]
\caption{Different fusion methods of avgpool and maxpool in Window enhanced attention on SAV.}
\label{fusion}
\centering
\begin{tabular*}{0.45\textwidth}{@{\extracolsep{\fill}}llll@{}}
\toprule 
Fusion way & mAP   & Params & FLOPs \\ \midrule
concat     & 67.51 & 172M   & 412G
\\
sum        & \textbf{67.90} & 172M   & 412G       \\ \bottomrule 
\end{tabular*}
\end{table}

\begin{figure}[!ht]
\centerline{\includegraphics[width=8cm]{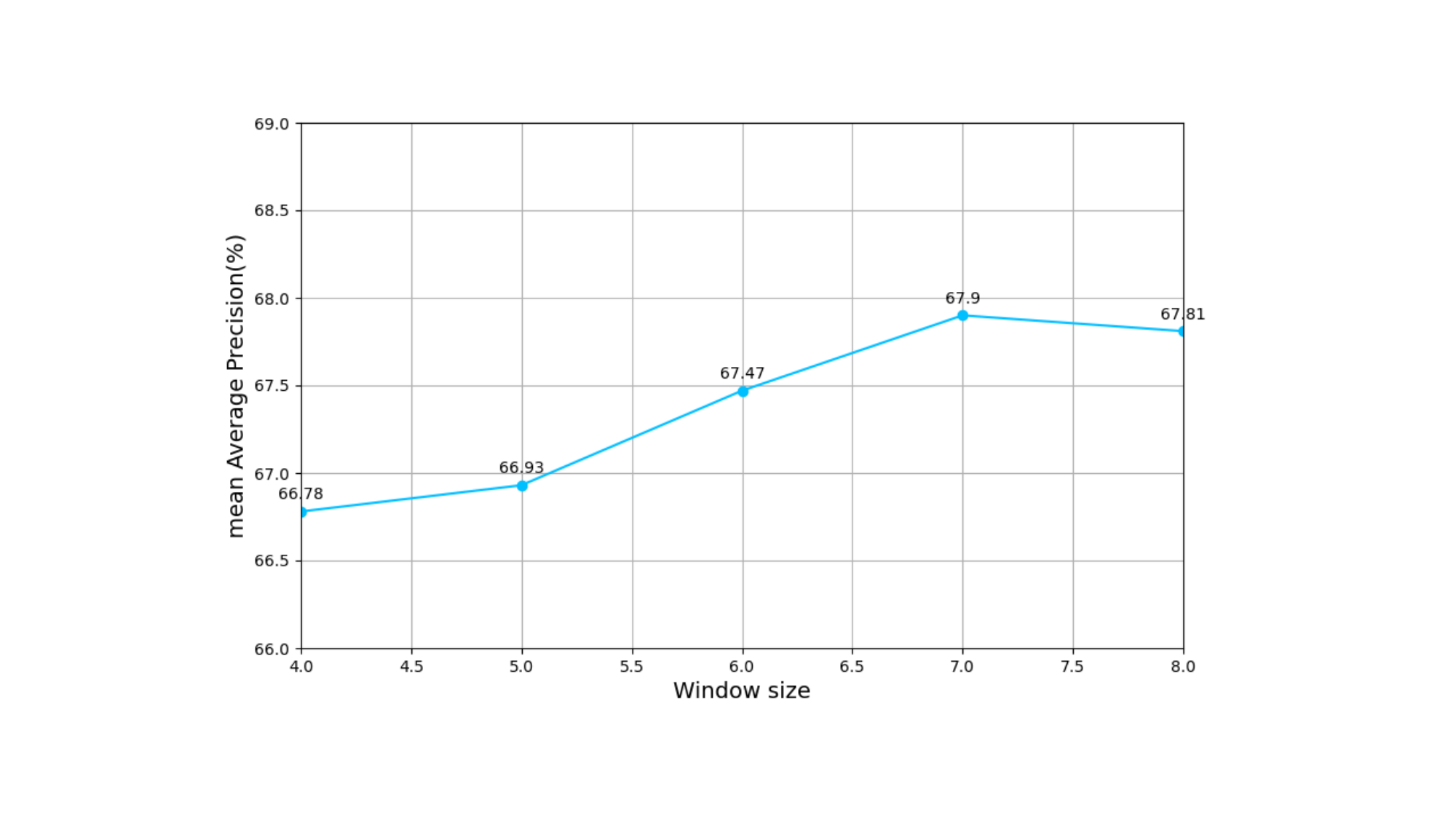}}
\caption{The impact of the window size on performance in the window enhanced attention module on SAV.}
\label{windowsize}
\end{figure}

\textbf{How much does the window size affect performance in the WEA?}
Fig. \ref{windowsize} explores the effect of window size on model performance in the WEA module. The horizontal coordinate indicates the window size and the vertical coordinate indicates the mAP. The results show that the model performance is gradually enhanced with increasing window size on SAV. The optimal performance is reached at a window size of 7, followed by a slow decrease. This suggests that a 7 × 7 window is sufficient to cover the key discriminative details of dense objects in SAV, such as eyes, mouth, and hands.

\begin{table}[]
\caption{Ablation for the Local Relation Aggregator and Window Enhanced Attention on SAV.}
\label{LRA AND WEA}
\centering
\begin{tabular*}{0.45\textwidth}{@{\extracolsep{\fill}}llll@{}}
\toprule 
Method                                                                                         & mAP   & Params & FLOPs \\ \midrule
ViT                                                                                            & 66.56 & 87M        & 180G      \\
+Local Relation Aggregator                                                                      & 67.79 &   144M     & 398G      \\
+Window Enhanced Attention                                                                     & 67.24 &   115M     &  194G     \\ \midrule
\begin{tabular}[c]{@{}l@{}}+Window Enhanced Attention\\ +Local Relation Aggregator\end{tabular} & \textbf{67.90} &   172M     & 412G      \\ \bottomrule 
\end{tabular*}
\end{table}

\textbf{How important are the LRA and WEA?}
Table \ref{LRA AND WEA} shows the ablation study on the LRA and WEA modules on SAV. These experiments use the ViT pre-trained by VideoMAE as the baseline method. First, the addition of the LRA to ViT leads to a 1.23\% improvement in mAP, with an increase of 57M in parameters and 218G in FLOPs. The LRA module dynamically encodes positional information through DWConv and learns local token affinities within small neighborhoods, thereby enhancing feature extraction early in the process and providing rich local details. While the use of MLPs and dense convolutional layers in LRA significantly increases computational cost, it effectively enhances the ability of the model to capture small interactions and dense relationships between objects, which is crucial for accurately recognizing actions in dense classroom scenes.
The addition of the WEA to the baseline results in a 0.68\% increase in mAP, accompanied by an additional 28M in parameters and 14G in FLOPs. Although the spatio-temporal attention mechanism in WEA adds some computational cost, it emphasizes the most discriminative regions within tokens, compensating for the shortcomings of the global self-attention mechanism that ignores small but critical details.
When both modules, the LRA and WEA, are used with ViT, the mAP is further increased by 1.34\% compared to the baseline. This improvement highlights the complementary strengths of the LRA and WEA in enhancing the ability of the model to perceive local dynamics and small objects in dense scenes. In future studies, it will be further investigated how to simplify the convolution operation in the LRA and the attention mechanism in the WEA to optimize the inference efficiency.

\subsection{Visualization results}

\begin{figure*}[!t]
\centerline{\includegraphics[width=18cm]{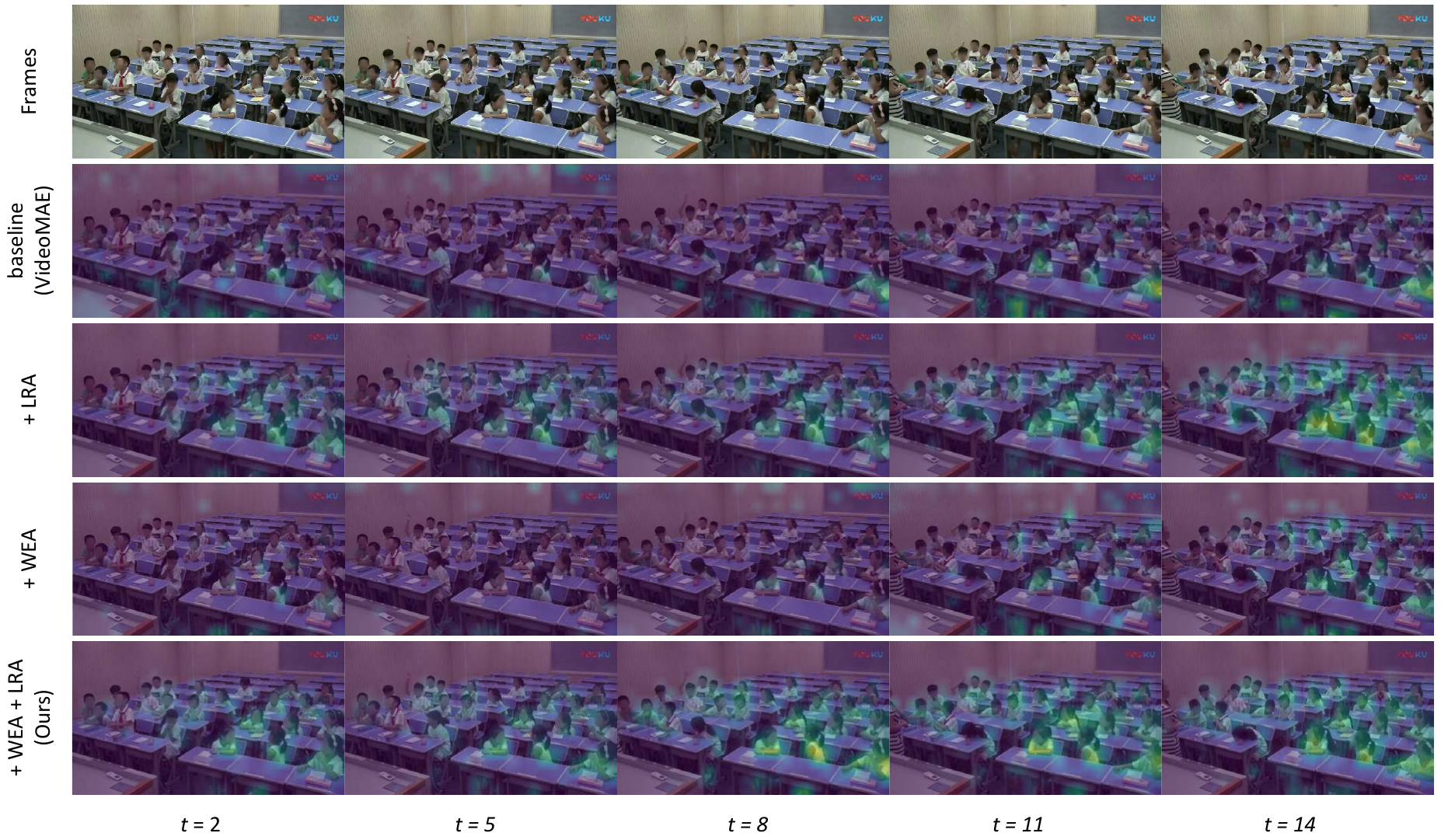}}
\caption{The Grad-Cam visualization of a sample test video clip generated from the baseline method (VideoMAE), the addition of LRA, the addition of WEA, and the addition of WEA and LRA to the baseline. Such robust qualitative results demonstrate the effectiveness of our method.}
\label{gradcam}
\end{figure*}

To better evaluate the performance of the model in dense and occluded environments, we use Grad-CAM \cite{selvaraju2020grad} to show the areas of the greatest concern that contribute the most to the ``sit” category in the sampled frames from the SAV dataset. As shown in Fig. \ref{gradcam}, our method can effectively focus on objects with scale differences in the front and back rows simultaneously, achieving more accurate attention coverage in occluded scenes. By utilizing the LRA to enhance fine-grained local interactions and the WEA to improve token-level attention, our method provides key complementary advantages for dense scenes with occlusions. In contrast, VideoMAE primarily focuses on larger and more prominent objects in the front row, ignoring smaller and occluded objects in the back row. Also, its visualization shows that its attention is dispersed to irrelevant background regions, indicating insufficient attention to occluded and dense objects. With the addition of LRA to the baseline, the focus on occluded objects in the back row is significantly improved. Since background distraction already exists in the baseline, the addition of the WEA module further exacerbates this distraction by refining the attention of the most responsive regions in each token, but also enhancing the attention to objects in the back row. Overall, our method provides robustness by reducing background attention distraction and increasing attention to dense and occluded objects in complex scenes.

\section{conclusion}

This paper introduces the SAV dataset, which contains 15 types of student actions in classroom scenes and poses new challenges to the field of action detection. The SAV dataset offers four key contributions: 1) It provides a variety of classroom scenarios from different courses and educational stages; 2) It challenges existing action detection models with issues such as subtle movements, dense objects, scale differences, high intra-class variability, and partial occlusion; 3) It necessitates the recognition of multiple actions occurring simultaneously among different participants in a multi-person situation; 4) It provides high-quality surveillance video data.
In addition, to benchmark this, we test several existing methods on the SAV dataset and propose a baseline method based on ViT. This method alleviates the problem that vanilla ViT has difficulty in capturing important local dynamics and discriminative tiny objects when facing small and dense objects in the SAV dataset. We introduce a Local Relation Aggregator and a Window Enhanced Attention module, which have been experimentally proven to outperform current representative methods. 
Future research will focus on the critical task of modeling more subtle and complex actions to fully interpret human activities in important educational contexts.

\bibliographystyle{IEEEtran}
\bibliography{references}

\begin{IEEEbiography}[{\includegraphics[width=1in,height=1.25in,clip,keepaspectratio]{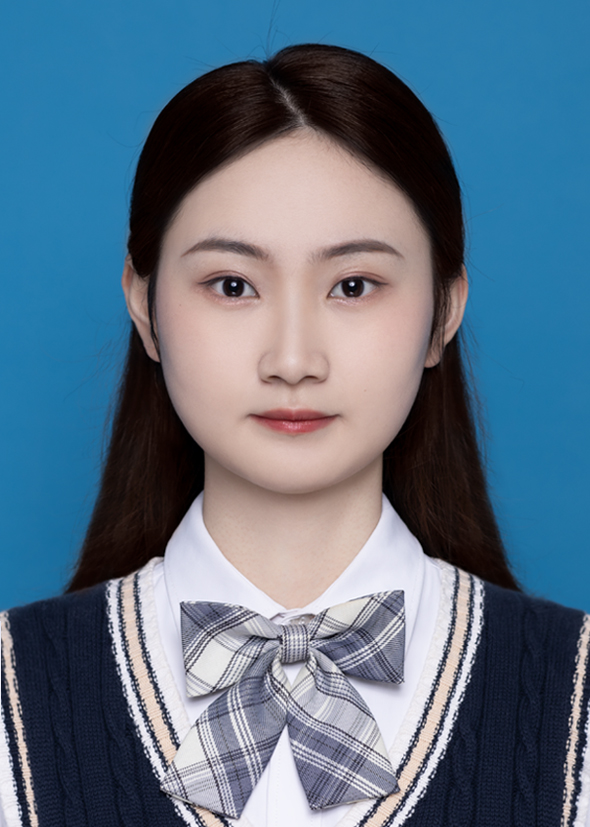}}]{Zhuolin Tan}
received the B.S. degree in electronic and information engineering from the Leshan Normal University, China, in 2020, and the M.S. degree from the Chongqing University of Posts and Telecommunications (CQUPT), Chongqing, China in 2023, where she is currently pursuing the Ph.D. degree. Her research interests include computer vision, image processing, deep learning, and video analysis.
\end{IEEEbiography}

\begin{IEEEbiography}[{\includegraphics[width=1in,height=1.25in,clip,keepaspectratio]{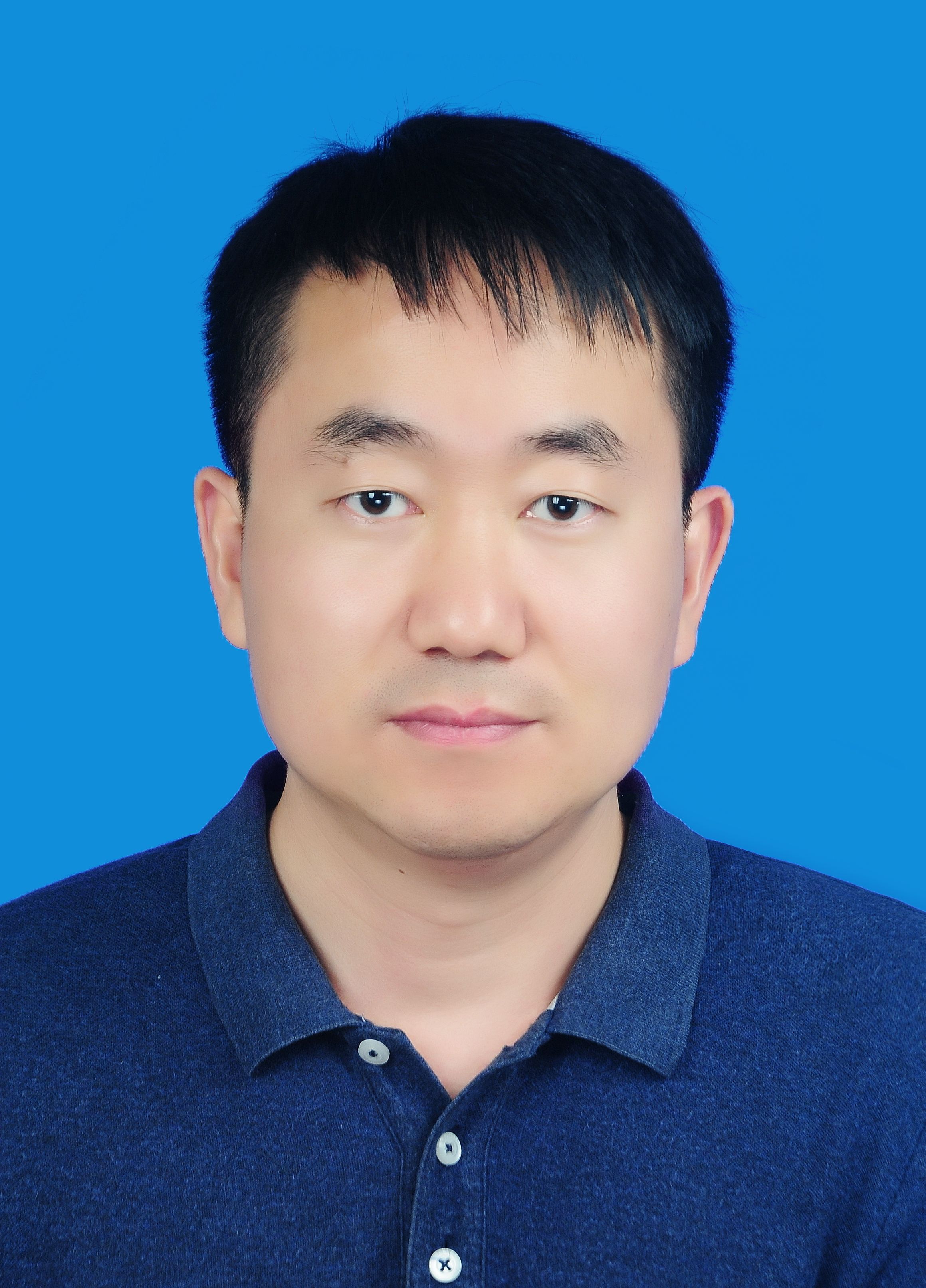}}]{Chenqiang Gao}
received the B.S. degree in computer science from the China University of Geosciences, Wuhan, China, in 2004 and the Ph.D. degree in control science and engineering from the Huazhong University of Science and Technology, Wuhan, China, in 2009. In August 2009, he joined the School of Communications and Information Engineering, Chongqing University of Posts and Telecommunications (CQUPT), Chongqing, China. In September 2012, he joined the Informedia Group with the School of Computer Science, Carnegie Mellon University, Pittsburgh, PA, USA, where he was a Visiting Scholar on multimedia event detection (MED) and surveillance event detection (SED). In April 2013, he became a Postdoctoral Fellow and continued work on MED and SED until March 2014, when he returned to CQUPT. In September 2023, he joined the School of Intelligent Systems Engineering, Sun Yat-sen University, Shenzhen, Guangdong, China. His research interests include image processing, infrared target detection, action recognition, and event detection. 
\end{IEEEbiography}

\begin{IEEEbiography}[{\includegraphics[width=1in,height=1.25in,clip,keepaspectratio]{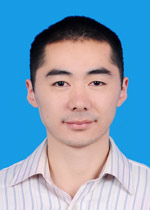}}]{Anyong Qin}
received the B.S. degree in information and computational science, and the Ph.D. degree in computer science from Chongqing University, Chongqing, China, in 2012, and 2019, respectively. He is currently a lecturer at the School of Communications and Information Engineering, Chongqing University of Posts and Telecommunications (CQUPT), Chongqing, China. His research interests include image processing, machine learning, and remote sensing images.
\end{IEEEbiography}

\begin{IEEEbiography}[{\includegraphics[width=1in,height=1.25in,clip,keepaspectratio]{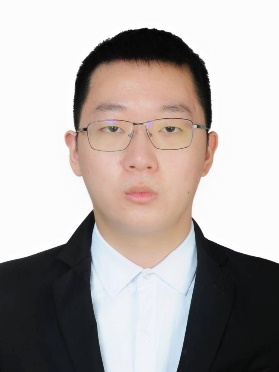}}]{Ruixin Chen}
received the B.S. degree from Chongqing University of Posts and Telecommunications, Chongqing, China, in 2022. He is currently pursuing the M.S. degree at the School of Communications and Information Engineering, Chongqing University of Posts and Telecommunications. His research interests include computer vision, video processing, and action recognition.
\end{IEEEbiography}

\begin{IEEEbiography}[{\includegraphics[width=1in,height=1.25in,clip,keepaspectratio]{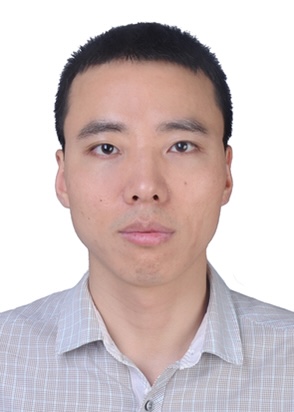}}]{Tiecheng Song}
received the Ph.D. degree in signal and information processing from the University of Electronic Science and Technology of China (UESTC), Chengdu, China, in 2015.  From October 2015 to April 2016, he was with the Multimedia Laboratory, Nanyang Technological University, Singapore, as a Visiting Student. He is currently a Professor with the School of Communications and Information Engineering, Chongqing University of Posts and Telecommunications (CQUPT), Chongqing, China.  His research interests include image processing and computer vision.
\end{IEEEbiography}

\begin{IEEEbiography}[{\includegraphics[width=1in,height=1.25in,clip,keepaspectratio]{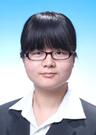}}]{Feng Yang}
received the B.S. degree in science and technology of remote sensing and the Ph.D. degree in photogrammetry and remote sensing from Wuhan University, Wuhan, China, in 2013 and 2018, respectively.
She is currently an Associate Professor with the School of Communications and Information Engineering, Chongqing University of Posts and Telecommunications (CQUPT), Chongqing, China. Her research interests include remote sensing imaging, video processing, and computer vision.
\end{IEEEbiography}

\begin{IEEEbiography}[{\includegraphics[width=1in,height=1.25in,clip,keepaspectratio]{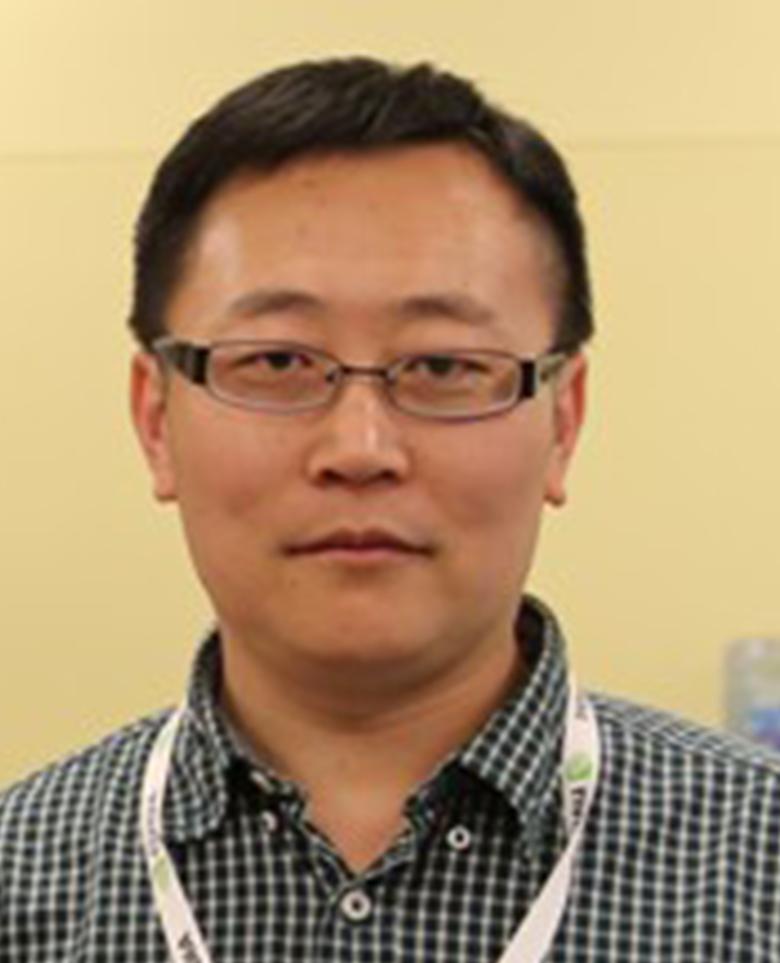}}]{Deyu Meng}
received the B.Sc., M.Sc., and Ph.D. degrees from Xi’an Jiaotong University, Xi’an, China, in 2001, 2004, and 2008, respectively. From 2012 to 2014, he took his two-year sabbatical leave with Carnegie Mellon University, Pittsburgh, PA, USA.  He is currently a Professor with the School of Mathematics and Statistics, Xi’an Jiaotong University, and an Adjunct Professor with the Faculty of Information Technology, Macau University of Science and Technology, Macau, China. His research interests include model-based deep learning, variational networks, and meta-learning.
\end{IEEEbiography}
\end{document}